\documentclass{article}
\usepackage[T1]{fontenc}
\usepackage{iclr2027_conference,times}
\usepackage{amssymb}

\usepackage{amsmath,amsfonts,bm}

\def\eqref#1{equation~\ref{#1}}

\def\1{\bm{1}}

\DeclareMathAlphabet{\mathsfit}{\encodingdefault}{\sfdefault}{m}{sl}
\SetMathAlphabet{\mathsfit}{bold}{\encodingdefault}{\sfdefault}{bx}{n}

\usepackage{graphicx}
\usepackage{float,wrapfig}
\usepackage{booktabs,tabularx,multirow,capt-of}
\usepackage{colortbl}
\usepackage{hyperref}
\definecolor{citationblue}{HTML}{315F86}
\usepackage{url}
\newcommand{\OURS}{EvoMoSkill}
\newcommand{\papertitle}{\OURS{}: Evolving a Mixture-of-Skills for LLM Agents}
\title{\raggedright\papertitle}
\author{%
  Kehua Feng, Yuqi Tang, Qiang Zhang, Keyan Ding$^{\ast}$,
  Huajun Chen\thanks{Corresponding authors.}\\
  {\normalfont Zhejiang University}\\
  {\normalfont\texttt{\{kehuafeng, dingkeyan\}@zju.edu.cn}}%
}
\iclrfinalcopy
\hypersetup{
  pdftitle={\papertitle},
  pdfauthor={Kehua Feng, Yuqi Tang, Qiang Zhang, Keyan Ding, Huajun Chen}
}
\begin{document}
\AddToShipoutPictureFG*{%
  \AtPageUpperLeft{%
    \put(\LenToUnit{108bp},\LenToUnit{-70.325bp}){%
      \includegraphics[height=19.55bp]{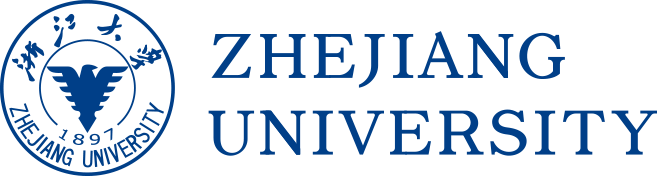}%
    }%
  }%
}
\maketitle
\lhead{\papertitle}

\begin{abstract}
Enabling agent skills to autonomously refine reusable knowledge is a central challenge in adapting large language model (LLM) agents to real-world applications. Recent work treats skills as trainable external parameters, following the iterative optimization paradigm of deep learning. However, these methods typically optimize a single skill across tasks, so revisions that benefit one task group can disrupt others. A skill library can limit this interference, but raises questions across skill creation, task routing, failure attribution, and evaluation. We introduce \OURS{}, a mixture-of-skills framework that jointly optimizes a share skill, expert skills, and a router. Each optimization step applies Credit-Assigned Gradient Descent (CreAGD) in five stages: (1) \textbf{Forward execution and credit assignment} attributes execution failures to skill text or router scope; (2) \textbf{Textual and routing gradient descent} edits skills, creates experts, and revises routing; (3) \textbf{Scope-conditioned paired evaluation} compares candidates with the original system on tasks they would serve; (4) \textbf{History-calibrated step acceptance} checks gains against past outcomes and accepts revisions after validation-set checks; and (5) \textbf{Optimizer memory accumulation} uses rejected drafts and feedback to improve later updates. Experiments with five LLMs on five benchmarks demonstrate state-of-the-art performance, substantially outperforming human-authored skills, LLM-generated skills, and baselines that optimize only a single skill. These results show that evolving mixture-of-skills yields more adaptable agents.
Code is available at \url{https://github.com/Irving-Feng/EvoMoSkill}.
\end{abstract}

\section{Introduction}
\label{sec:introduction}

Recent advances in large language models (LLMs) are expanding the scope of autonomous problem solving, enabling agents to coordinate multistep planning, tool execution, and environmental feedback within increasingly complex workflows~\citep{wang2023voyager,li2026skillsbench}. Agent skills consolidate reusable procedures into external documents, enabling adaptation without model parameter updates~\citep{zhang2025agentskills,jiang2026sokskills}. However, manual authoring and one-shot generation alone cannot incorporate lessons from subsequent executions. Enabling autonomous skill self-improvement as task demands evolve therefore remains a central challenge.

A promising approach treats skills as trainable external parameters, drawing on the iterative optimization paradigm of deep learning. Batch execution and scoring provide the analogue of a forward pass, while an optimizer derives textual update directions from trajectory errors to guide skill edits~\citep{liu2026skillforge}. Evaluation selects revisions for subsequent batches~\citep{alzubi2026evoskill}. These methods typically train a single skill shared across tasks~\citep{yang2026skillopt,lin2026skillboost}. Every revision therefore changes the guidance supplied to all tasks, even when their procedural requirements differ. A repair for one group can consequently disrupt behavior that already works elsewhere. A library of separately selectable and editable skills can localize such adaptations, but raises questions beyond the single-skill training loop: \emph{When should new skills be introduced, and which tasks should use them? How should failures be attributed to skills or routing decisions, and what constitutes an improvement to the library as a whole?}

\begin{figure}[h]
 \vspace{-2em}
 \centering
 \includegraphics[width=\linewidth]{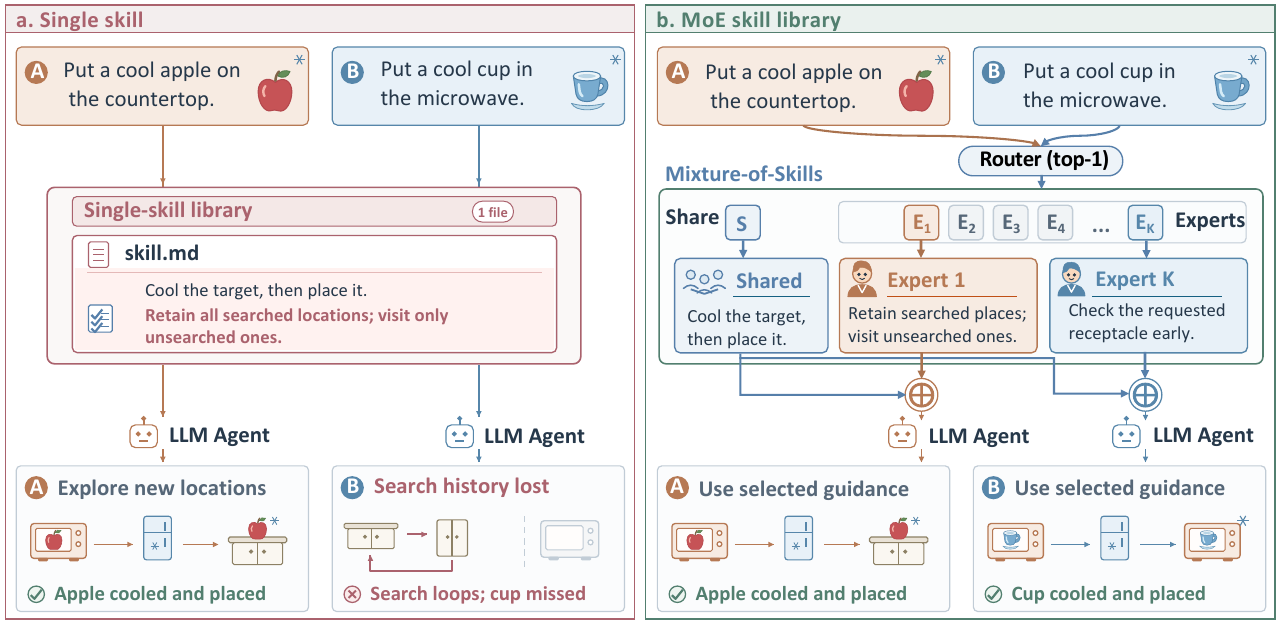}
 \caption{Different tasks can require different skill guidance. (a) A single search rule helps task A explore and place the apple, but task B loses its search history, revisits locations, and misses the cup. (b) The router combines the share skill with distinct expert skills: task A retains searched locations, while task B checks the requested receptacle early. Both tasks succeed.}
 \label{fig:motivation}
\end{figure}

To this end, we propose \OURS{}, a mixture-of-skills framework that jointly optimizes a share skill, expert skills, and their router (Figure~\ref{fig:motivation}). Training tasks are organized into families of related problems, providing a common basis for revision and evaluation. Each optimization step applies Credit-Assigned Gradient Descent (CreAGD) in five stages. (1) \textbf{Forward execution and credit assignment} maps failures to responsible skills or router scopes. (2) \textbf{Textual and routing gradient descent} proposes bounded edits, new experts, and scope changes. (3) \textbf{Scope-conditioned paired evaluation} compares candidate skills with the prior system on tasks they would serve. (4) \textbf{History-calibrated step acceptance} evaluates gains across repeated runs against historical performance, then validates promising revisions on validation tasks. (5) \textbf{Optimizer memory accumulation} uses rejected drafts and feedback to guide later revisions. Accepted experts join the library and improve through subsequent feedback. Persistently declining experts retire, making room for new candidates.

Across five diverse agent benchmarks, \OURS{} improves average performance by 14.9--23.8 percentage points over no skill for five student models. Ablation studies further attribute these gains to individual components and support the overall design of \OURS{}.

Our contributions are threefold:
\begin{itemize}
 \item We formulate skill evolution as joint optimization of skill content and applicability. \OURS{} evolves a share skill, expert skills tailored to task families, and their routing scopes, adapting both the agent's procedures and where each procedure applies.
 \item We introduce Credit-Assigned Gradient Descent (CreAGD), which attributes execution errors to skills or routing, generates multiple targeted edits, and accepts skill updates only after deliberative validation of gains and cross-task regressions.
 \item Across five benchmarks and five students, \OURS{} outperforms the strongest baseline by 4.0--6.8 points on average. Ablations confirm the value of routing and guarded updates.
\end{itemize}

\section{Related Work}
\label{sec:related_work}

\paragraph{Agent Skills as Reusable Knowledge.}
An agent skill is a reusable package of procedural knowledge, typically organized as a \texttt{SKILL.md} document with optional scripts and references. It guides task planning, tool use, and result verification~\citep{jiang2026sokskills,zhang2025agentskills,destefanis2026gitskills}. Human authors translate domain expertise into instructions agents can follow~\citep{destefanis2026authoring}. To accelerate authoring, automated skill generation systems translate requirements into skills, as in \texttt{Skill Creator}~\citep{anthropic2026skillcreator}, or construct them from documents and scientific resources~\citep{si2026ctx2skill,shen2026skillfoundry}. With this design, systems consolidate experience into libraries, retrieving, composing, and selectively loading skills~\citep{liang2026skillnet,li2026agentskillos,zheng2026skillrouter,gao2026skillreducer} for document processing, software development, scientific analysis, and embodied interaction~\citep{li2026skillsbench,han2026sweskills,ju2026embodiskill,wang2023voyager}. However, manual authoring and one-shot generation alone cannot accumulate lessons from subsequent executions~\citep{ni2026trace2skill,yang2026skillopt}.

\paragraph{Evolution of Agentic Skills.}
Skill evolution continually revises reusable procedures through execution experience~\citep{ding2026survey,zhong2026skilllearnbench}. Feedback determines the direction of evolution. \emph{Execution signals} use task scores, runtime errors, and verification results to localize failures~\citep{wang2023voyager,ma2026skillgen,tian2026skillscoach}. \emph{Trajectory comparisons} contrast successful and failed runs or align teacher demonstrations with student failures to isolate missing behaviors~\citep{chen2026skillcat,shi2026skillkd}. \emph{User and model judgments} draw on dialogue preferences, critiques, and simulated interactions to expose knowledge and procedural gaps~\citep{yang2026autoskill,zhang2026coevoskills,shinn2023reflexion,madaan2023selfrefine,yan2026skillevo}. Turning feedback into lasting gains requires deciding how to update skills, reuse them, and verify improvements in task performance. \emph{Model and skill coevolution} uses task rewards to train models for skill selection, execution, or curation, while new rollouts drive skill creation and refinement~\citep{xia2026skillrl,tu2026d2skill,shi2026skill1,ouyang2026skillos,zhang2026memskill}. \emph{Skill-library evolution} distills reusable procedures from trajectories, consolidates redundant knowledge, and revises stored skills~\citep{zhao2024expel,wang2024awm,ni2026trace2skill,jiang2026xskill,wang2026skillx}, with retrieval and routing governing reuse~\citep{zhou2026mementoskills}. Candidate search evaluates which edits to retain~\citep{alzubi2026evoskill,yuksekgonul2025textgrad,agrawal2026gepa}. Within this direction, \emph{stepwise skill optimization} repeatedly executes training tasks, diagnoses shared failures, and revises skills across batches or epochs~\citep{liu2026skillforge,yang2026skillopt,lin2026skillboost}. Analogous to supervised training, successive data-driven updates refine an external skill, with edit budgets, validation checks, or regression constraints controlling revisions. This approach accumulates experience across tasks and supports measurable improvements without retraining the executor, producing reusable domain skills. Yet single-skill optimizers such as SkillOpt and SkillBoost apply one skill throughout a domain, where heterogeneous task families may require incompatible procedures~\citep{li2026skillsbench,han2026sweskills}. \OURS{} extends stepwise optimization to a mixture-of-skills library, jointly evolving the share skill, expert skills, and routing scopes. 

\section{Method}
\label{sec:method}
\begin{figure}[t]
 \centering
 \vspace{-1.5em}
 \includegraphics[width=\linewidth]{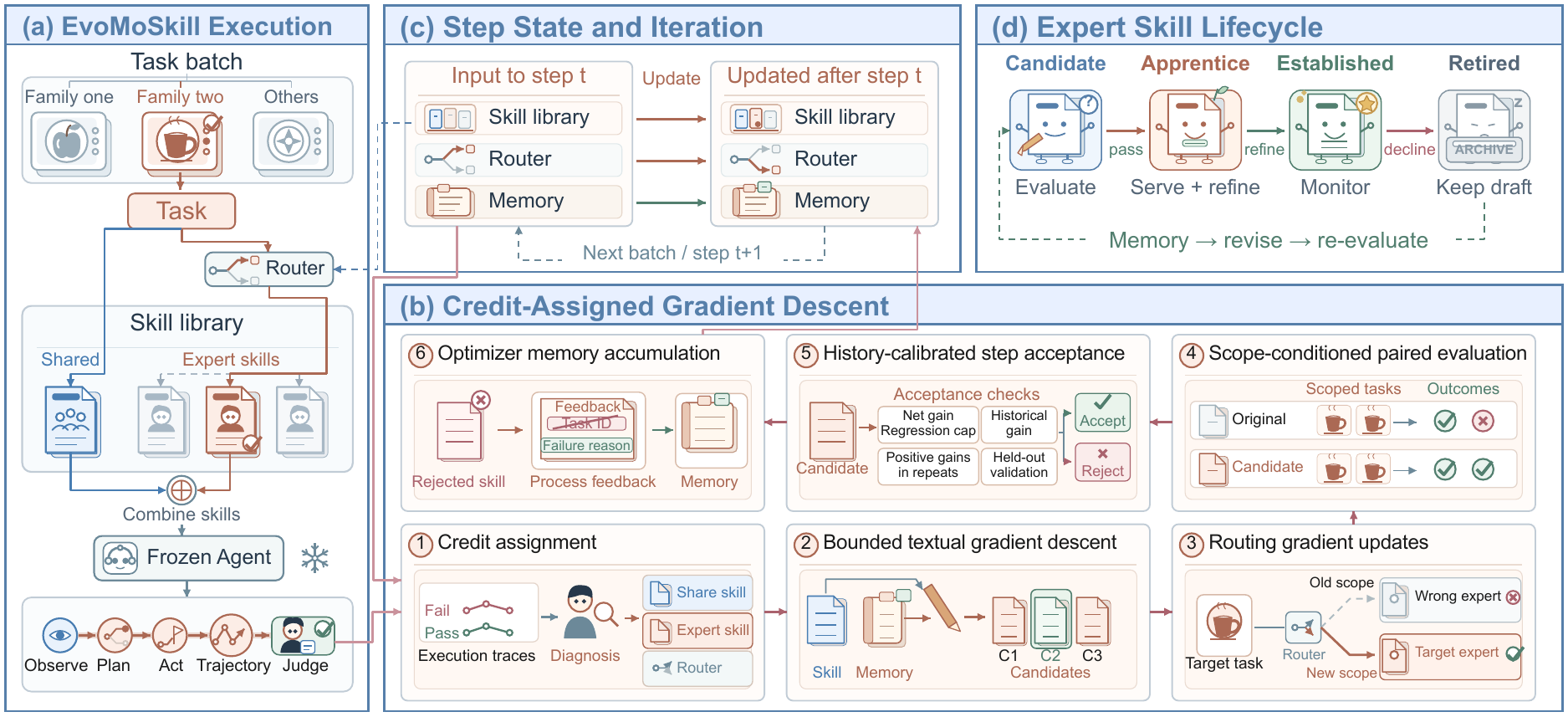}\vspace{-0.5em}
 \caption{Overview of \OURS{}. Tasks use the share skill and at most one routed expert skill. CreAGD assigns failure feedback, proposes skill revisions, and selects them through repeated paired evaluation and validation. Routing errors trigger scope repair. Memory guides future proposals, and successive updates govern expert skill admission, refinement, and retirement.}
 \label{fig:overview}
 \vspace{-0.5em}
\end{figure}

In this section, we introduce \OURS{}, a skill optimization framework that jointly optimizes a mixture-of-skills library and router from execution feedback. As shown in Figure~\ref{fig:overview}, \OURS{} (1) organizes training batches by task family, (2) uses Credit-Assigned Gradient Descent (CreAGD) to assign failure feedback, propose skill and routing revisions, and evaluate skill updates, and (3) refines the share skill and manages expert skill lifecycles across optimization steps.

\subsection{Problem Formulation}
\label{sec:formulation}
\label{sec:specialization}
\OURS{} extends skill training inspired by deep learning~\citep{yang2026skillopt,lin2026skillboost} to jointly optimize a library $\mathcal L=(s_{\mathrm{sh}},\{s_j\}_{j\in I})$ and router $g$, with a frozen executor $M$. The share skill $s_{\mathrm{sh}}$ provides general guidance, while expert skills $s_j$ supply specialized procedures, with $|I|\leq K_{\max}$. The LLM router matches task descriptions to expert scopes and returns $g(x;\mathcal L)\in I\cup\{\varnothing\}$, where $s_{\varnothing}$ denotes empty context.
For task $x$, $M$ samples trajectory $\xi$ from the task and skill context, and $\mathrm{Eval}$ returns a binary signal $y$ indicating whether the task was completed successfully:
\begin{equation}
 \xi\sim M(\,\cdot\mid x\oplus s_{\mathrm{sh}}\oplus s_{g(x;\mathcal L)}),\qquad
 y=\mathrm{Eval}(x,\xi)\in\{0,1\}.
 \label{eq:mixture}
\end{equation}
Here, $\oplus$ denotes context concatenation.
Given disjoint splits $\mathcal D_{\mathrm{train}}$, $\mathcal D_{\mathrm{val}}$, and $\mathcal D_{\mathrm{test}}$, our objective is to jointly learn the library and router that maximize mean validation success:
\begin{equation}
 \begin{aligned}
 (\mathcal L^\star,g^\star)&\in\arg\max_{(\mathcal L,g)\in\mathcal P(\mathcal D_{\mathrm{train}})} J_{\mathcal D_{\mathrm{val}}}(\mathcal L,g),\\
 J_{\mathcal D}(\mathcal L,g)&=\frac{1}{|\mathcal D|}\sum_{x\in\mathcal D}\mathbb E[y\mid x,\mathcal L,g].
 \end{aligned}
 \label{eq:objective}
\end{equation}
Here, $\mathcal P(\mathcal D_{\mathrm{train}})$ contains library and router pairs obtained during training. $J_{\mathcal D}$ averages task success over sampled trajectories. We report $J_{\mathcal D_{\mathrm{test}}}(\mathcal L^\star,g^\star)$ for the frozen pair on test tasks.

\subsection{Training Data and Batches}
\label{sec:overview}
We partition $\mathcal D_{\mathrm{train}}$ into $F$ task families $\{\mathcal D_f\}_{f=1}^{F}$ using task-type annotations or, when unavailable, semantic clustering of task descriptions. We merge related families to keep $F\leq K_{\max}$ and reserve one expert skill per family. Each epoch comprises multiple optimization steps. At step $t$, we sample a batch $\mathcal D_{\mathrm{batch},t}$ across all families, denoting its family-$f$ tasks by $\mathcal D_{\mathrm{batch},f,t}$. Tasks execute with the previous library $\mathcal L_{t-1}$ and router $g_{t-1}$. Family labels also partition validation tasks into $\mathcal D_{\mathrm{val},f}$ for skill evaluation and library selection. The test set $\mathcal D_{\mathrm{test}}$ remains unavailable during training and library selection and is used only for final evaluation.

\subsection{Credit-Assigned Gradient Descent}
\label{sec:optimization}
Credit-Assigned Gradient Descent (CreAGD) assigns textual gradients to responsible skills or the router at each step. An LLM optimizer uses the assigned gradients and stored feedback to propose revisions, which undergo paired evaluation before acceptance.

\begin{wraptable}{r}{0.50\textwidth}
 \centering
 \scriptsize
 \makeatletter
 \vspace{-2em}\caption{Failure diagnoses and update targets.}\vspace{0.5em}
 \label{tab:attribution}
 \begingroup
 \fontsize{6.2pt}{7.5pt}\selectfont %
 \renewcommand{\arraystretch}{1.15}
 \setlength{\tabcolsep}{1.5pt}
 \begin{tabularx}{\linewidth}{>{\raggedright\arraybackslash}p{0.17\linewidth}>{\raggedright\arraybackslash}X>{\raggedright\arraybackslash}p{0.22\linewidth}}
 \toprule
 \rowcolor{blue!5} Diagnosis & Evidence from execution & Update target\\
 \midrule
 Coverage & No dedicated expert skill covers the family. & New expert skill\\
 Misguide & Following a skill rule contributes to failure. & Skill content\\
 Missing & The skill omits an action needed for the task. & Skill content\\
 Routing & The selected expert skill does not fit the task. & Router scope\\
 Capability & Correct actions fail or the cause is uncertain. & Defer revision\\
 \bottomrule
 \end{tabularx}
 \endgroup
\end{wraptable}
\paragraph{Forward execution and credit assignment.}
The frozen executor $M$ runs each $x_i\in\mathcal D_{\mathrm{batch},t}$ for $R\geq2$ repetitions, producing trajectories $\{\xi_{i,r}\}_{r=1}^{R}$ and binary outcomes $\{y_{i,r}\}_{r=1}^{R}$.

A more capable teacher LLM compares failed trajectories from $M$ with successful demonstrations on the same tasks and examines the applicable skill rules~\citep{shi2026skillkd}. Table~\ref{tab:attribution} maps the resulting diagnoses to updates. Missing or misleading instructions are assigned to the share skill when failures span families, or to an expert skill when confined to one family. A family without coverage calls for a new expert, while selection of an unsuitable expert calls for a router-scope revision. Intermittent failures can motivate reliability edits. If appropriate actions still fail or the cause is unclear, the teacher defers revision and reassesses the case when new trajectories become available.
Each actionable diagnosis is assigned to an editable component $u\in\mathcal U_t$. The value $u=\mathrm{sh}$ denotes share-skill text, and $u=j\in I$ denotes the text of existing expert $s_j$. A proposed expert receives a fresh index $u=\vert I\vert+1$, while $u=\mathrm{route}$ denotes router scope. Diagnoses assigned to the same component form its textual gradient:
\begin{equation}
 G_{u,t}=\operatorname{Aggregate}\!\left(\{d_{i,t}:a(d_{i,t})=u\}\right).
 \label{eq:credit}
\end{equation}
Here $d_{i,t}$ is the diagnostic evidence for task $x_i$, and $a$ assigns it to the responsible skill text or router scope. Deferred cases receive no update target.

\paragraph{Bounded textual gradient descent.}
For each skill-text target $u\in\mathcal U_t\setminus\{\mathrm{route}\}$, the gradient identifies failures to correct and successful behavior to retain. The optimizer proposes $N$ drafts $\{\tilde s_{u,t}^{(c)}\}_{c=1}^{N}$ from different perspectives. For an existing skill, each draft applies at most $B$ local edits:
\begin{equation}
 \begin{aligned}
 E_{u,t}^{(c)}&\sim\operatorname{Optimizer}(G_{u,t},s_{u,t-1},\mathcal M_{u,t-1}),
 &|E_{u,t}^{(c)}|&\leq B,\\
 \tilde s_{u,t}^{(c)}&=\operatorname{Apply}(s_{u,t-1},E_{u,t}^{(c)}),
 &c&=1,\ldots,N.
 \end{aligned}
 \label{eq:descent}
\end{equation}
Here $N$ is the number of drafts per skill-text target at each step. Each $E_{u,t}^{(c)}$ contains insertions, deletions, or replacements. Memory $\mathcal M_{u,t-1}$ supplies past drafts and feedback for the same skill text, helping avoid repeated failures. New experts are drafted in full without the edit bound. We discard instance-specific rules and duplicate procedures. Share skill candidates contain cross-family guidance, while expert candidates contain family-specific procedures. Batch screening on $\mathcal D_{\mathrm{batch},t}$ selects one draft $\tilde s_{u,t}$ per skill-text target for paired evaluation.

\paragraph{Routing gradient descent.}
For $u=\mathrm{route}$, $G_{\mathrm{route},t}$ revises expert scopes without changing skill text. The new scope enters $g_t$ and is tested at the next step without extra rollouts. For a new expert, the optimizer drafts a scope alongside the skill text and adds both to a temporary library $\tilde{\mathcal L}_t$ and router $\tilde g_t$. Before admission, we check that the scope routes enough target-family tasks to the new expert without routing checked tasks from other families to it. If either check fails, we revise the scope and repeat the checks.

\paragraph{Scope-conditioned paired evaluation.}
We compare each candidate system with the unchanged original system $(\mathcal L_{t-1},g_{t-1})$ on a paired-evaluation set $Q$. For share-skill edits, $Q=\mathcal D_{\mathrm{batch},t}$. For an existing expert assigned to family $f$, $Q=\mathcal D_{\mathrm{batch},f,t}$. The candidate system replaces only that expert text in $\mathcal L_{t-1}$, preserving all other skill texts and the original router $g_{t-1}$.

For a new expert assigned to family $f$, the temporary router $\tilde g_t$ selects $Q$ from $\mathcal D_{\mathrm{batch},f,t}$ by retaining only tasks routed to that expert. On $Q$, the candidate system uses the share skill and new expert in $\tilde{\mathcal L}_t$, while the original system uses its share skill and any expert selected by $g_{t-1}$ in $\mathcal L_{t-1}$. The preceding scope check separately tests tasks from other families.

Each system performs $R$ runs on every task in $Q$, yielding original outcomes $\{y_{i,r}\}_{r=1}^{R}$ and candidate outcomes $\{\tilde y_{i,r}\}_{r=1}^{R}$. The net gain $\Delta_Q$ counts additional successes. The stable regression count $q_Q$ counts tasks that succeed in all original runs but fail in all candidate runs:
\begin{equation}
 \Delta_Q=\sum_{x_i\in Q}\sum_{r=1}^{R}(\tilde y_{i,r}-y_{i,r}),\qquad
 q_Q=\sum_{x_i\in Q}\mathbf1\!\left[\sum_r y_{i,r}=R\ \land\ \sum_r\tilde y_{i,r}=0\right].
 \label{eq:gain}
\end{equation}
\paragraph{History-calibrated step acceptance.}
Expert skill candidates first undergo three checks. \textbf{(1) Current gain and regression:} we require $\Delta_Q\geq\tau>0$ and $q_Q\leq\rho_{\mathrm{exp}}$. \textbf{(2) Historical calibration:} we guard against an unusually weak current baseline. Past successes $H_i$ and executions $N_i$ on $x_i$ under unchanged skills and routing give $p_i^{\mathrm{hist}}=H_i/N_i$. Replacing current baseline counts with $Rp_i^{\mathrm{hist}}$ expected successes yields $\Delta_Q^{\mathrm{hist}}$; both gains must reach $\tau$. For tasks without history, we use the current baseline rate. Large discrepancies between current and historical rates trigger additional executions.
\textbf{(3) Rollout consistency:} at least $\eta R$ repetitions must yield positive gains across $Q$. Writing the gain in repetition $r$ as $\Delta_Q^{(r)}$, the three checks become:
\begin{equation}
 \begin{gathered}
 \Delta_Q^{\mathrm{hist}}=\sum_{x_i\in Q}\left(\sum_{r=1}^{R}\tilde y_{i,r}-Rp_i^{\mathrm{hist}}\right),\qquad
 \Delta_Q^{(r)}=\sum_{x_i\in Q}(\tilde y_{i,r}-y_{i,r}),\\
 \min\{\Delta_Q,\Delta_Q^{\mathrm{hist}}\}\geq\tau,\quad q_Q\leq\rho_{\mathrm{exp}},\quad
 \frac1R\sum_{r=1}^{R}\mathbf1[\Delta_Q^{(r)}>0]\geq\eta.
 \end{gathered}
 \label{eq:acceptance}
\end{equation}
Share skill updates pass training acceptance when $q_Q\leq\rho_{\mathrm{sh}}$ on the full batch, restricting persistent harm across task families without imposing a positive gain threshold.

Passing candidates then undergo \textbf{(4) Held-out validation:} paired tests on $\mathcal D_{\mathrm{val},f}$ (experts) or $\mathcal D_{\mathrm{val}}$ (share skill). Excessive stable regressions reject both types, while a substantial success loss also rejects experts. Accepted updates and scope repairs yield the next library $\mathcal L_t$ and router $g_t$.

\paragraph{Optimizer memory accumulation.}
\label{par:optimizer_memory}
Memory $\mathcal M_{u,t}$ stores rejected drafts, outcomes, and rejection reasons for each target $u$. Records remove task identifiers and replace entities with roles, preserving errors such as missing prerequisites. The optimizer retrieves these records when proposing revisions for the same target, using past outcomes to refine drafts and avoid repeating unsuccessful repairs.

\subsection{Skill Library Lifecycle}
\label{sec:population}
Starting with an empty library $\mathcal L_0$, CreAGD drafts the first skills from execution failures. Once introduced, the share skill remains active throughout training and accumulates cross-family guidance through revisions that pass global evaluation. Newly drafted experts remain \textit{candidates} while their text and scope are evaluated. Those that pass training and validation acceptance enter $\mathcal L_t$ as \textit{apprentices} and begin serving tasks at step $t+1$. Their subsequent execution feedback guides further CreAGD revisions to skill text and routing scope, enabling them to become \textit{established} experts.

We monitor each expert on fixed tasks, comparing current and admission success rates over the same number of executions. Sustained deterioration beyond a tolerance triggers \textit{retirement}. Retired skills remain inactive \textit{work-in-progress drafts} in optimizer memory for later revision and reevaluation. Repeated admission failures require a cooldown before retrying.



\section{Experiments}
\label{sec:experiments}
\subsection{Experimental Setup}
\label{sec:experimental_setup}

\paragraph{Benchmarks.}
We evaluate \OURS{} on five complementary benchmarks. ALFWorld~\citep{shridhar2021alfworld} tests household task execution in text environments, while ScienceWorld~\citep{wang2022scienceworld} requires interactive scientific experimentation. SearchQA~\citep{dunn2017searchqa} evaluates question answering with retrieved snippets, SpreadsheetBench~\citep{ma2024spreadsheetbench} tests real-world workbook manipulation, and SkillLearnBench~\citep{zhong2026skilllearnbench} evaluates learning reusable procedures for real-world tasks. We use disjoint training, validation, and test splits. ALFWorld, SpreadsheetBench, and SkillLearnBench use task success, ScienceWorld uses its official best score, and SearchQA uses exact match.

\paragraph{Baselines.}
We compare six baselines. No skill runs the student without skill guidance, following~\citet{yang2026skillopt}. LLM Skill uses GPT-5.6-Sol to create one skill document with Skill Creator~\citep{anthropic2026skillcreator} from each benchmark's task description, example tasks, and answer trajectories. It does not revise the document. Trace2Skill~\citep{ni2026trace2skill} consolidates lessons extracted from individual trajectories into reusable skills. SkillOpt~\citep{yang2026skillopt} makes bounded skill-text edits from batch feedback and selects updates through validation. EvoSkill~\citep{alzubi2026evoskill} analyzes failures to propose and select skill revisions. SKILL-KD~\citep{shi2026skillkd} contrasts teacher and student trajectories, tests proposed patches through student reruns, and consolidates the resulting rules.

\paragraph{Models and protocol.}
For \OURS{} and every baseline, we test all five benchmarks with five frozen students: DeepSeek-V4-Flash~\citep{deepseek2026v4flash}, Claude-Opus-4.6~\citep{anthropic2026opus46}, Kimi-K2.6~\citep{moonshot2026kimik26}, Qwen3-32B~\citep{qwen2025qwen332b}, and Qwen3.5-4B~\citep{qwen2026qwen354b}. In our controlled comparison, all skill-learning methods use GPT-5.6-Sol~\citep{openai2026gpt56} as the common teacher and skill optimizer. The ReAct execution harness, tools, test tasks, and scorer remain fixed for each benchmark and student. Further experimental settings appear in Appendix~\ref{app:experiments}.

\subsection{Main Results}
\label{sec:main_results}

Table~\ref{tab:main_results} reports each benchmark result and the average across benchmarks. On unseen ALFWorld and ScienceWorld tasks, \OURS{} achieves the highest score for all five students. It leads outright in 23 of 25 benchmark--student comparisons and improves the average by 14.9--23.8 percentage points over No skill and by 4.0--6.8 points over the strongest competing baseline for each student. These gains across distinct task types show the benefit of evolving a selectively routed skill library.

\begin{table}[t]
 \vspace{-2em}
 \centering
 \setlength{\belowcaptionskip}{5pt}
 \caption{Main results on five benchmarks across five student models (\%). Metrics are task success, SearchQA exact match, and ScienceWorld official best score. Green/red subscripts show percentage-point change from No skill. Bold/underline mark the best/second-best result.}
 \label{tab:main_results}
 \begingroup
 \footnotesize
 \setlength{\tabcolsep}{4.5pt}
 \renewcommand{\arraystretch}{1.12}
 \newcommand{\resultcell}[3]{\makebox[2.6em][r]{#1}\makebox[2.9em][l]{\raisebox{-0.25ex}{\scriptsize\color{#2}#3}}}
 \newcommand{\basedcell}[1]{\makebox[2.6em][r]{#1}\makebox[2.9em][l]{}}
 \resizebox{\linewidth}{!}{%
 \begin{tabular}{@{}ll*{6}{c}@{}}
 \toprule
 \textbf{Model} & \textbf{Method} & \textbf{ALFWorld} & \textbf{SciWorld} & \textbf{SearchQA} & \textbf{Spreadsheet} & \textbf{SkillLearn} & \textbf{Avg.}\\
 \midrule
  & No skill & \basedcell{58.2} & \basedcell{61.7} & \basedcell{77.6} & \basedcell{49.6} & \basedcell{15.4} & \basedcell{52.5}\\
  & LLM Skill & \resultcell{59.7}{green!45!black}{+1.5} & \resultcell{53.0}{red!75!black}{-8.7} & \resultcell{73.2}{red!75!black}{-4.4} & \resultcell{46.8}{red!75!black}{-2.9} & \resultcell{15.4}{green!45!black}{+0.0} & \resultcell{49.6}{red!75!black}{-2.9}\\
  & SkillOpt & \resultcell{76.9}{green!45!black}{+18.7} & \resultcell{\underline{71.1}}{green!45!black}{+9.4} & \resultcell{82.8}{green!45!black}{+5.2} & \resultcell{67.5}{green!45!black}{+17.9} & \resultcell{23.1}{green!45!black}{+7.7} & \resultcell{64.3}{green!45!black}{+11.8}\\
 DeepSeek-V4-F. & Trace2Skill & \resultcell{64.2}{green!45!black}{+6.0} & \resultcell{65.8}{green!45!black}{+4.0} & \resultcell{75.9}{red!75!black}{-1.7} & \resultcell{69.3}{green!45!black}{+19.7} & \resultcell{19.2}{green!45!black}{+3.8} & \resultcell{58.9}{green!45!black}{+6.4}\\
  & EvoSkill & \resultcell{78.4}{green!45!black}{+20.2} & \resultcell{54.4}{red!75!black}{-7.4} & \resultcell{\underline{83.5}}{green!45!black}{+5.9} & \resultcell{55.0}{green!45!black}{+5.4} & \resultcell{\underline{26.9}}{green!45!black}{+11.5} & \resultcell{59.6}{green!45!black}{+7.1}\\
  & SKILL-KD & \resultcell{\underline{82.1}}{green!45!black}{+23.9} & \resultcell{61.7}{green!45!black}{+0.0} & \resultcell{83.1}{green!45!black}{+5.4} & \resultcell{\underline{75.4}}{green!45!black}{+25.7} & \resultcell{\underline{26.9}}{green!45!black}{+11.5} & \resultcell{\underline{65.8}}{green!45!black}{+13.3}\\
 \rowcolor{blue!7}\cellcolor{white}  & \textbf{\OURS{}} & \resultcell{\textbf{90.3}}{green!45!black}{+32.1} & \resultcell{\textbf{78.5}}{green!45!black}{+16.8} & \resultcell{\textbf{84.4}}{green!45!black}{+6.7} & \resultcell{\textbf{79.3}}{green!45!black}{+29.7} & \resultcell{\textbf{30.8}}{green!45!black}{+15.4} & \resultcell{\textbf{72.6}}{green!45!black}{+20.1}\\
 \midrule
  & No skill & \basedcell{80.6} & \basedcell{75.2} & \basedcell{82.8} & \basedcell{56.8} & \basedcell{15.4} & \basedcell{62.1}\\
  & LLM Skill & \resultcell{84.3}{green!45!black}{+3.7} & \resultcell{73.2}{red!75!black}{-2.0} & \resultcell{83.0}{green!45!black}{+0.2} & \resultcell{53.9}{red!75!black}{-2.9} & \resultcell{11.5}{red!75!black}{-3.8} & \resultcell{61.2}{red!75!black}{-1.0}\\
  & SkillOpt & \resultcell{88.8}{green!45!black}{+8.2} & \resultcell{81.2}{green!45!black}{+6.0} & \resultcell{\underline{86.9}}{green!45!black}{+4.1} & \resultcell{71.8}{green!45!black}{+15.0} & \resultcell{\underline{26.9}}{green!45!black}{+11.5} & \resultcell{71.1}{green!45!black}{+9.0}\\
 Claude-Opus-4.6 & Trace2Skill & \resultcell{85.8}{green!45!black}{+5.2} & \resultcell{78.5}{green!45!black}{+3.4} & \resultcell{84.8}{green!45!black}{+2.0} & \resultcell{67.1}{green!45!black}{+10.4} & \resultcell{19.2}{green!45!black}{+3.8} & \resultcell{67.1}{green!45!black}{+5.0}\\
  & EvoSkill & \resultcell{87.3}{green!45!black}{+6.7} & \resultcell{81.9}{green!45!black}{+6.7} & \resultcell{84.1}{green!45!black}{+1.3} & \resultcell{66.1}{green!45!black}{+9.3} & \resultcell{15.4}{green!45!black}{+0.0} & \resultcell{66.9}{green!45!black}{+4.8}\\
  & SKILL-KD & \resultcell{\underline{90.3}}{green!45!black}{+9.7} & \resultcell{\underline{82.6}}{green!45!black}{+7.4} & \resultcell{85.8}{green!45!black}{+3.0} & \resultcell{\underline{78.2}}{green!45!black}{+21.4} & \resultcell{23.1}{green!45!black}{+7.7} & \resultcell{\underline{72.0}}{green!45!black}{+9.8}\\
 \rowcolor{blue!7}\cellcolor{white}  & \textbf{\OURS{}} & \resultcell{\textbf{95.5}}{green!45!black}{+14.9} & \resultcell{\textbf{89.9}}{green!45!black}{+14.8} & \resultcell{\textbf{88.1}}{green!45!black}{+5.4} & \resultcell{\textbf{80.7}}{green!45!black}{+23.9} & \resultcell{\textbf{30.8}}{green!45!black}{+15.4} & \resultcell{\textbf{77.0}}{green!45!black}{+14.9}\\
 \midrule
  & No skill & \basedcell{61.2} & \basedcell{66.4} & \basedcell{80.7} & \basedcell{32.5} & \basedcell{7.7} & \basedcell{49.7}\\
  & LLM Skill & \resultcell{69.4}{green!45!black}{+8.2} & \resultcell{66.4}{green!45!black}{+0.0} & \resultcell{80.4}{red!75!black}{-0.4} & \resultcell{38.6}{green!45!black}{+6.1} & \resultcell{7.7}{green!45!black}{+0.0} & \resultcell{52.5}{green!45!black}{+2.8}\\
  & SkillOpt & \resultcell{79.9}{green!45!black}{+18.7} & \resultcell{73.8}{green!45!black}{+7.4} & \resultcell{\textbf{85.6}}{green!45!black}{+4.9} & \resultcell{\underline{47.5}}{green!45!black}{+15.0} & \resultcell{19.2}{green!45!black}{+11.5} & \resultcell{61.2}{green!45!black}{+11.5}\\
 Kimi-K2.6 & Trace2Skill & \resultcell{67.2}{green!45!black}{+6.0} & \resultcell{65.8}{red!75!black}{-0.7} & \resultcell{80.2}{red!75!black}{-0.5} & \resultcell{38.9}{green!45!black}{+6.4} & \resultcell{11.5}{green!45!black}{+3.8} & \resultcell{52.7}{green!45!black}{+3.0}\\
  & EvoSkill & \resultcell{74.6}{green!45!black}{+13.4} & \resultcell{68.5}{green!45!black}{+2.0} & \resultcell{83.1}{green!45!black}{+2.4} & \resultcell{41.8}{green!45!black}{+9.3} & \resultcell{15.4}{green!45!black}{+7.7} & \resultcell{56.7}{green!45!black}{+7.0}\\
  & SKILL-KD & \resultcell{\underline{81.3}}{green!45!black}{+20.2} & \resultcell{\underline{78.5}}{green!45!black}{+12.1} & \resultcell{83.2}{green!45!black}{+2.5} & \resultcell{45.4}{green!45!black}{+12.9} & \resultcell{\underline{23.1}}{green!45!black}{+15.4} & \resultcell{\underline{62.3}}{green!45!black}{+12.6}\\
 \rowcolor{blue!7}\cellcolor{white}  & \textbf{\OURS{}} & \resultcell{\textbf{90.3}}{green!45!black}{+29.1} & \resultcell{\textbf{84.6}}{green!45!black}{+18.1} & \resultcell{\underline{84.6}}{green!45!black}{+3.9} & \resultcell{\textbf{48.6}}{green!45!black}{+16.1} & \resultcell{\textbf{26.9}}{green!45!black}{+19.2} & \resultcell{\textbf{67.0}}{green!45!black}{+17.3}\\
 \midrule
  & No skill & \basedcell{40.3} & \basedcell{30.9} & \basedcell{72.3} & \basedcell{22.5} & \basedcell{0.0} & \basedcell{33.2}\\
  & LLM Skill & \resultcell{59.7}{green!45!black}{+19.4} & \resultcell{26.9}{red!75!black}{-4.0} & \resultcell{69.6}{red!75!black}{-2.7} & \resultcell{25.0}{green!45!black}{+2.5} & \resultcell{0.0}{green!45!black}{+0.0} & \resultcell{36.2}{green!45!black}{+3.0}\\
  & SkillOpt & \resultcell{76.1}{green!45!black}{+35.8} & \resultcell{\underline{46.3}}{green!45!black}{+15.4} & \resultcell{\underline{80.1}}{green!45!black}{+7.8} & \resultcell{39.6}{green!45!black}{+17.1} & \resultcell{\underline{15.4}}{green!45!black}{+15.4} & \resultcell{\underline{51.5}}{green!45!black}{+18.3}\\
 Qwen3-32B & Trace2Skill & \resultcell{64.2}{green!45!black}{+23.9} & \resultcell{31.5}{green!45!black}{+0.7} & \resultcell{72.1}{red!75!black}{-0.2} & \resultcell{24.6}{green!45!black}{+2.1} & \resultcell{3.8}{green!45!black}{+3.8} & \resultcell{39.3}{green!45!black}{+6.1}\\
  & EvoSkill & \resultcell{72.4}{green!45!black}{+32.1} & \resultcell{40.3}{green!45!black}{+9.4} & \resultcell{76.1}{green!45!black}{+3.8} & \resultcell{31.1}{green!45!black}{+8.6} & \resultcell{7.7}{green!45!black}{+7.7} & \resultcell{45.5}{green!45!black}{+12.3}\\
  & SKILL-KD & \resultcell{\underline{82.8}}{green!45!black}{+42.5} & \resultcell{43.0}{green!45!black}{+12.1} & \resultcell{78.7}{green!45!black}{+6.4} & \resultcell{\underline{40.4}}{green!45!black}{+17.9} & \resultcell{11.5}{green!45!black}{+11.5} & \resultcell{51.3}{green!45!black}{+18.1}\\
 \rowcolor{blue!7}\cellcolor{white}  & \textbf{\OURS{}} & \resultcell{\textbf{86.6}}{green!45!black}{+46.3} & \resultcell{\textbf{47.7}}{green!45!black}{+16.8} & \resultcell{\textbf{81.0}}{green!45!black}{+8.7} & \resultcell{\textbf{42.9}}{green!45!black}{+20.4} & \resultcell{\textbf{19.2}}{green!45!black}{+19.2} & \resultcell{\textbf{55.5}}{green!45!black}{+22.3}\\
 \midrule
  & No skill & \basedcell{32.1} & \basedcell{22.2} & \basedcell{68.2} & \basedcell{9.6} & \basedcell{0.0} & \basedcell{26.4}\\
  & LLM Skill & \resultcell{30.6}{red!75!black}{-1.5} & \resultcell{27.5}{green!45!black}{+5.4} & \resultcell{64.9}{red!75!black}{-3.4} & \resultcell{13.6}{green!45!black}{+3.9} & \resultcell{0.0}{green!45!black}{+0.0} & \resultcell{27.3}{green!45!black}{+0.9}\\
  & SkillOpt & \resultcell{\underline{77.6}}{green!45!black}{+45.5} & \resultcell{\underline{36.9}}{green!45!black}{+14.8} & \resultcell{74.3}{green!45!black}{+6.1} & \resultcell{14.3}{green!45!black}{+4.7} & \resultcell{\underline{7.7}}{green!45!black}{+7.7} & \resultcell{42.2}{green!45!black}{+15.7}\\
 Qwen3.5-4B & Trace2Skill & \resultcell{54.5}{green!45!black}{+22.4} & \resultcell{28.2}{green!45!black}{+6.0} & \resultcell{61.9}{red!75!black}{-6.3} & \resultcell{11.1}{green!45!black}{+1.4} & \resultcell{\textbf{11.5}}{green!45!black}{+11.5} & \resultcell{33.4}{green!45!black}{+7.0}\\
  & EvoSkill & \resultcell{66.4}{green!45!black}{+34.3} & \resultcell{34.2}{green!45!black}{+12.1} & \resultcell{68.6}{green!45!black}{+0.4} & \resultcell{19.3}{green!45!black}{+9.7} & \resultcell{3.8}{green!45!black}{+3.8} & \resultcell{38.5}{green!45!black}{+12.1}\\
  & SKILL-KD & \resultcell{70.2}{green!45!black}{+38.1} & \resultcell{34.9}{green!45!black}{+12.8} & \resultcell{\underline{78.8}}{green!45!black}{+10.6} & \resultcell{\underline{23.9}}{green!45!black}{+14.3} & \resultcell{\textbf{11.5}}{green!45!black}{+11.5} & \resultcell{\underline{43.9}}{green!45!black}{+17.4}\\
 \rowcolor{blue!7}\cellcolor{white}  & \textbf{\OURS{}} & \resultcell{\textbf{88.1}}{green!45!black}{+56.0} & \resultcell{\textbf{41.6}}{green!45!black}{+19.5} & \resultcell{\textbf{81.1}}{green!45!black}{+12.9} & \resultcell{\textbf{28.9}}{green!45!black}{+19.3} & \resultcell{\textbf{11.5}}{green!45!black}{+11.5} & \resultcell{\textbf{50.3}}{green!45!black}{+23.8}\\
 \bottomrule
 \end{tabular}%
 }
 \endgroup
 \vspace{-0.5em}
\end{table}

The one-shot LLM Skill baseline is less reliable: relative to No skill, it improves 9 of 25 comparisons, reduces performance in 11, and leaves 5 unchanged. On SearchQA, four of five students perform worse with that skill. By contrast, \OURS{} improves on No skill in all 25 comparisons. For Qwen3.5-4B, ALFWorld rises from 32.1\% to 88.1\% and SpreadsheetBench from 9.6\% to 28.9\%. The contrast supports execution-guided skill revision and selective reuse as a more reliable route to gains across tasks and students with different no-skill performance.

\begin{figure}[t]
 \centering
 \begin{minipage}[t]{0.588\linewidth}
  \vspace{0pt}
  \includegraphics[width=0.49\linewidth]{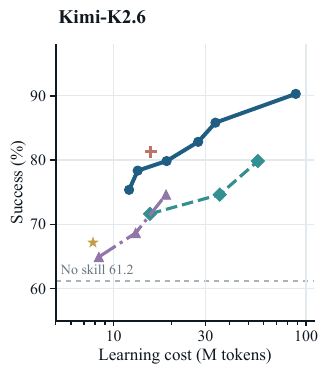}\hfill
  \includegraphics[width=0.49\linewidth]{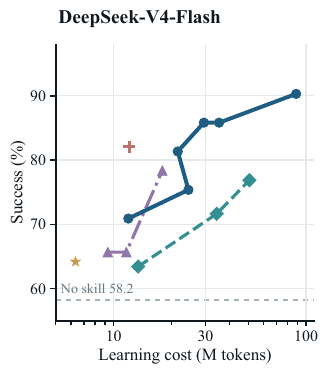}
  \vspace{-0.35em}
  \includegraphics[width=\linewidth]{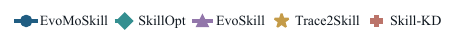}
  \caption{ALFWorld unseen success versus learning cost (M tokens). Gray dashed lines show No skill.}
  \label{fig:alfworld_budget_frontier}
 \end{minipage}\hfill
 \begin{minipage}[t]{0.390\linewidth}
  \vspace{0pt}
  \centering
  \begingroup
\fontsize{6.15pt}{6.75pt}\selectfont
\setlength{\tabcolsep}{1.4pt}
\renewcommand{\arraystretch}{1.0}
\setlength{\aboverulesep}{1.0pt}
\setlength{\belowrulesep}{1.0pt}
\begin{tabular*}{\linewidth}{@{\extracolsep{\fill}}lllrr@{}}
\toprule
\textbf{Benchmark} & \textbf{Source} & \textbf{Target} & \textbf{LLM} & \textbf{Transfer} \\
\midrule
\multirow{6}{*}{ALFWorld} & \multirow{2}{*}{Qwen3-32B} & Qwen3.5-4B & 30.6 & 82.1\raisebox{-0.25ex}{\fontsize{4.8pt}{5.4pt}\selectfont\textcolor{green!45!black}{+51.5}} \\
 &  & Opus-4.6 & 84.3 & 87.3\raisebox{-0.25ex}{\fontsize{4.8pt}{5.4pt}\selectfont\textcolor{green!45!black}{+3.0}} \\
\cmidrule(lr){2-5}
 & \multirow{2}{*}{Qwen3.5-4B} & Qwen3-32B & 59.7 & 76.9\raisebox{-0.25ex}{\fontsize{4.8pt}{5.4pt}\selectfont\textcolor{green!45!black}{+17.2}} \\
 &  & Opus-4.6 & 84.3 & 91.0\raisebox{-0.25ex}{\fontsize{4.8pt}{5.4pt}\selectfont\textcolor{green!45!black}{+6.7}} \\
\cmidrule(lr){2-5}
 & \multirow{2}{*}{Opus-4.6} & Qwen3-32B & 59.7 & 80.6\raisebox{-0.25ex}{\fontsize{4.8pt}{5.4pt}\selectfont\textcolor{green!45!black}{+20.9}} \\
 &  & Qwen3.5-4B & 30.6 & 64.9\raisebox{-0.25ex}{\fontsize{4.8pt}{5.4pt}\selectfont\textcolor{green!45!black}{+34.3}} \\
\midrule
\multirow{6}{*}{SearchQA} & \multirow{2}{*}{Qwen3-32B} & Qwen3.5-4B & 64.9 & 77.4\raisebox{-0.25ex}{\fontsize{4.8pt}{5.4pt}\selectfont\textcolor{green!45!black}{+12.6}} \\
 &  & Opus-4.6 & 83.0 & 86.9\raisebox{-0.25ex}{\fontsize{4.8pt}{5.4pt}\selectfont\textcolor{green!45!black}{+3.9}} \\
\cmidrule(lr){2-5}
 & \multirow{2}{*}{Qwen3.5-4B} & Qwen3-32B & 69.6 & 78.1\raisebox{-0.25ex}{\fontsize{4.8pt}{5.4pt}\selectfont\textcolor{green!45!black}{+8.5}} \\
 &  & Opus-4.6 & 83.0 & 87.4\raisebox{-0.25ex}{\fontsize{4.8pt}{5.4pt}\selectfont\textcolor{green!45!black}{+4.4}} \\
\cmidrule(lr){2-5}
 & \multirow{2}{*}{Opus-4.6} & Qwen3-32B & 69.6 & 75.1\raisebox{-0.25ex}{\fontsize{4.8pt}{5.4pt}\selectfont\textcolor{green!45!black}{+5.5}} \\
 &  & Qwen3.5-4B & 64.9 & 72.4\raisebox{-0.25ex}{\fontsize{4.8pt}{5.4pt}\selectfont\textcolor{green!45!black}{+7.6}} \\
\midrule
\multirow{6}{*}{\shortstack[l]{Spreadsheet\\Bench}} & \multirow{2}{*}{Qwen3-32B} & Qwen3.5-4B & 13.6 & 27.5\raisebox{-0.25ex}{\fontsize{4.8pt}{5.4pt}\selectfont\textcolor{green!45!black}{+13.9}} \\
 &  & Opus-4.6 & 53.9 & 74.3\raisebox{-0.25ex}{\fontsize{4.8pt}{5.4pt}\selectfont\textcolor{green!45!black}{+20.4}} \\
\cmidrule(lr){2-5}
 & \multirow{2}{*}{Qwen3.5-4B} & Qwen3-32B & 25.0 & 36.4\raisebox{-0.25ex}{\fontsize{4.8pt}{5.4pt}\selectfont\textcolor{green!45!black}{+11.4}} \\
 &  & Opus-4.6 & 53.9 & 69.3\raisebox{-0.25ex}{\fontsize{4.8pt}{5.4pt}\selectfont\textcolor{green!45!black}{+15.4}} \\
\cmidrule(lr){2-5}
 & \multirow{2}{*}{Opus-4.6} & Qwen3-32B & 25.0 & 39.3\raisebox{-0.25ex}{\fontsize{4.8pt}{5.4pt}\selectfont\textcolor{green!45!black}{+14.3}} \\
 &  & Qwen3.5-4B & 13.6 & 23.6\raisebox{-0.25ex}{\fontsize{4.8pt}{5.4pt}\selectfont\textcolor{green!45!black}{+10.0}} \\
\bottomrule
\end{tabular*}
\endgroup
\vspace{-0.2em}
  \captionof{table}{Cross-model transfer (\%). Green subscripts: gains vs. LLM Skill.}
  \label{tab:main_transfer}
 \end{minipage}
\end{figure}

\textbf{Learning cost and benchmark performance.} On ALFWorld, \OURS{} improves across learning budgets for both students and attains the highest final success (Figure~\ref{fig:alfworld_budget_frontier}). Later checkpoints outperform other iterative skill methods at comparable or lower cost. Continued library optimization thus turns additional learning budget into stronger unseen-task performance.

\textbf{Cross-model transfer.} We evaluate frozen libraries on students other than those used for training (Table~\ref{tab:main_transfer}). All 18 transfers across three benchmarks exceed both No skill and target-model LLM Skill. Gains across every source--target pair show that learned skill texts and routing scopes remain useful with a different executor, extending the benefit of optimization beyond the training student.

\subsection{Ablation Studies: Why Does the Mixture-of-Skills Help?}
\label{sec:mechanistic_results}

\paragraph{Q1: Does the mixture-of-skills organization improve performance?}
We train three alternatives from scratch. \textit{Single skill} optimizes one document for all tasks. \textit{Experts only} learns routed expert skills without a share skill. \textit{Static partition} trains a separate skill for each task family and routes tasks by their descriptions without updating those assignments. \OURS{} exceeds the first two in all nine benchmark--student settings and Static partition in seven (Table~\ref{tab:ablation_library}). On ALFWorld and SearchQA, it leads every alternative for all three students; on \mbox{SpreadsheetBench}, it leads Single skill and Experts only for all three. These results favor joint optimization of the share skill and routed experts over simpler library designs across the tested tasks and students.


\begin{table}[!ht]
 \centering
 \setlength{\belowcaptionskip}{3pt}
 \caption{Skill-library organization ablation. Black values are benchmark scores (\%). Gray subscripts show percentage-point changes from \OURS{}. Blue/red shading marks gains/losses.}
 \label{tab:ablation_library}
 \begingroup
 \footnotesize
 \setlength{\tabcolsep}{2.5pt}
 \renewcommand{\arraystretch}{1.12}
 \resizebox{\linewidth}{!}{%
 \begin{tabular}{@{}l*{9}{c}@{}}
 \toprule
 & \multicolumn{3}{c}{\textbf{ALFWorld}} & \multicolumn{3}{c}{\textbf{SearchQA}} & \multicolumn{3}{c}{\textbf{SpreadsheetBench}} \\
 \cmidrule(lr){2-4}\cmidrule(lr){5-7}\cmidrule(l){8-10}
 \textbf{Method} & \textbf{Qwen3-32B} & \textbf{Qwen3.5-4B} & \textbf{Opus-4.6} & \textbf{Qwen3-32B} & \textbf{Qwen3.5-4B} & \textbf{Opus-4.6} & \textbf{Qwen3-32B} & \textbf{Qwen3.5-4B} & \textbf{Opus-4.6} \\
 \midrule
 \textbf{\OURS{}} & \makebox[2.3em][r]{\textbf{86.6}}\makebox[2.2em][l]{} & \makebox[2.3em][r]{\textbf{88.1}}\makebox[2.2em][l]{} & \makebox[2.3em][r]{\textbf{95.5}}\makebox[2.2em][l]{} & \makebox[2.3em][r]{\textbf{81.0}}\makebox[2.2em][l]{} & \makebox[2.3em][r]{\textbf{81.1}}\makebox[2.2em][l]{} & \makebox[2.3em][r]{\textbf{88.1}}\makebox[2.2em][l]{} & \makebox[2.3em][r]{\textbf{42.9}}\makebox[2.2em][l]{} & \makebox[2.3em][r]{\textbf{28.9}}\makebox[2.2em][l]{} & \makebox[2.3em][r]{\textbf{80.7}}\makebox[2.2em][l]{} \\
 Single skill & \cellcolor[HTML]{E8B4A9}\makebox[2.3em][r]{\textcolor{black}{76.1}}\makebox[2.2em][l]{\raisebox{-0.25ex}{\scriptsize\textcolor{black!55}{-10.5}}} & \cellcolor[HTML]{EABBB1}\makebox[2.3em][r]{\textcolor{black}{79.1}}\makebox[2.2em][l]{\raisebox{-0.25ex}{\scriptsize\textcolor{black!55}{-9.0}}} & \cellcolor[HTML]{F0D1C9}\makebox[2.3em][r]{\textcolor{black}{89.6}}\makebox[2.2em][l]{\raisebox{-0.25ex}{\scriptsize\textcolor{black!55}{-6.0}}} & \cellcolor[HTML]{FAF3F0}\makebox[2.3em][r]{\textcolor{black}{79.9}}\makebox[2.2em][l]{\raisebox{-0.25ex}{\scriptsize\textcolor{black!55}{-1.1}}} & \cellcolor[HTML]{EDC5BC}\makebox[2.3em][r]{\textcolor{black}{73.6}}\makebox[2.2em][l]{\raisebox{-0.25ex}{\scriptsize\textcolor{black!55}{-7.6}}} & \cellcolor[HTML]{F6E7E1}\makebox[2.3em][r]{\textcolor{black}{85.3}}\makebox[2.2em][l]{\raisebox{-0.25ex}{\scriptsize\textcolor{black!55}{-2.9}}} & \cellcolor[HTML]{F1D5CE}\makebox[2.3em][r]{\textcolor{black}{37.5}}\makebox[2.2em][l]{\raisebox{-0.25ex}{\scriptsize\textcolor{black!55}{-5.4}}} & \cellcolor[HTML]{E8B4A9}\makebox[2.3em][r]{\textcolor{black}{18.2}}\makebox[2.2em][l]{\raisebox{-0.25ex}{\scriptsize\textcolor{black!55}{-10.7}}} & \cellcolor[HTML]{ECC1B7}\makebox[2.3em][r]{\textcolor{black}{72.5}}\makebox[2.2em][l]{\raisebox{-0.25ex}{\scriptsize\textcolor{black!55}{-8.2}}} \\
 Experts only & \cellcolor[HTML]{ECC1B7}\makebox[2.3em][r]{\textcolor{black}{78.4}}\makebox[2.2em][l]{\raisebox{-0.25ex}{\scriptsize\textcolor{black!55}{-8.2}}} & \cellcolor[HTML]{EFCBC3}\makebox[2.3em][r]{\textcolor{black}{81.3}}\makebox[2.2em][l]{\raisebox{-0.25ex}{\scriptsize\textcolor{black!55}{-6.7}}} & \cellcolor[HTML]{F8EBE6}\makebox[2.3em][r]{\textcolor{black}{93.3}}\makebox[2.2em][l]{\raisebox{-0.25ex}{\scriptsize\textcolor{black!55}{-2.2}}} & \cellcolor[HTML]{F8ECE7}\makebox[2.3em][r]{\textcolor{black}{78.9}}\makebox[2.2em][l]{\raisebox{-0.25ex}{\scriptsize\textcolor{black!55}{-2.1}}} & \cellcolor[HTML]{F8EDE8}\makebox[2.3em][r]{\textcolor{black}{79.1}}\makebox[2.2em][l]{\raisebox{-0.25ex}{\scriptsize\textcolor{black!55}{-2.0}}} & \cellcolor[HTML]{F7EAE5}\makebox[2.3em][r]{\textcolor{black}{85.7}}\makebox[2.2em][l]{\raisebox{-0.25ex}{\scriptsize\textcolor{black!55}{-2.4}}} & \cellcolor[HTML]{ECC1B7}\makebox[2.3em][r]{\textcolor{black}{34.6}}\makebox[2.2em][l]{\raisebox{-0.25ex}{\scriptsize\textcolor{black!55}{-8.2}}} & \cellcolor[HTML]{F0D0C8}\makebox[2.3em][r]{\textcolor{black}{22.9}}\makebox[2.2em][l]{\raisebox{-0.25ex}{\scriptsize\textcolor{black!55}{-6.1}}} & \cellcolor[HTML]{F3DDD6}\makebox[2.3em][r]{\textcolor{black}{76.4}}\makebox[2.2em][l]{\raisebox{-0.25ex}{\scriptsize\textcolor{black!55}{-4.3}}} \\
 Static partition & \cellcolor[HTML]{F6E6E0}\makebox[2.3em][r]{\textcolor{black}{83.6}}\makebox[2.2em][l]{\raisebox{-0.25ex}{\scriptsize\textcolor{black!55}{-3.0}}} & \cellcolor[HTML]{F2D6CF}\makebox[2.3em][r]{\textcolor{black}{82.8}}\makebox[2.2em][l]{\raisebox{-0.25ex}{\scriptsize\textcolor{black!55}{-5.2}}} & \cellcolor[HTML]{FBF6F2}\makebox[2.3em][r]{\textcolor{black}{94.8}}\makebox[2.2em][l]{\raisebox{-0.25ex}{\scriptsize\textcolor{black!55}{-0.7}}} & \cellcolor[HTML]{FAF4F1}\makebox[2.3em][r]{\textcolor{black}{80.1}}\makebox[2.2em][l]{\raisebox{-0.25ex}{\scriptsize\textcolor{black!55}{-0.9}}} & \cellcolor[HTML]{F9F0EC}\makebox[2.3em][r]{\textcolor{black}{79.6}}\makebox[2.2em][l]{\raisebox{-0.25ex}{\scriptsize\textcolor{black!55}{-1.6}}} & \cellcolor[HTML]{F9F2EE}\makebox[2.3em][r]{\textcolor{black}{86.9}}\makebox[2.2em][l]{\raisebox{-0.25ex}{\scriptsize\textcolor{black!55}{-1.3}}} & \cellcolor[HTML]{F6F8F7}\makebox[2.3em][r]{\textcolor{black}{43.9}}\makebox[2.2em][l]{\raisebox{-0.25ex}{\scriptsize\textcolor{black!55}{+1.1}}} & \cellcolor[HTML]{F6E4DF}\makebox[2.3em][r]{\textcolor{black}{25.7}}\makebox[2.2em][l]{\raisebox{-0.25ex}{\scriptsize\textcolor{black!55}{-3.2}}} & \cellcolor[HTML]{F8F9F7}\makebox[2.3em][r]{\textcolor{black}{81.4}}\makebox[2.2em][l]{\raisebox{-0.25ex}{\scriptsize\textcolor{black!55}{+0.7}}} \\
 \bottomrule
 \end{tabular}%
 }
 \endgroup
\end{table}

\paragraph{Q2: Which acceptance checks make skill updates reliable?}
We ablate the four acceptance checks of Section~\ref{sec:optimization} in turn: \textit{rollout consistency}, \textit{historical calibration}, \textit{the regression bound}, and \textit{held-out validation}. Removing the regression bound lowers test-set scores in all nine benchmark--student settings by 6.4 points on average (Table~\ref{tab:ablation_acceptance}). Weakening or removing each other check lowers scores in eight of nine settings. Together, these results indicate that repeated trials and historical calibration guard against chance gains, while regression and validation checks screen harmful edits. The checks therefore make updates more reliable across tasks and students.


\begin{table}[H]
 \centering
 \setlength{\belowcaptionskip}{3pt}
 \caption{Update acceptance ablation. Black values are benchmark scores (\%). Gray subscripts show percentage-point changes from \OURS{}. Blue/red shading marks gains/losses.}
 \label{tab:ablation_acceptance}
 \begingroup
 \footnotesize
 \setlength{\tabcolsep}{2.5pt}
 \renewcommand{\arraystretch}{1.12}
 \resizebox{\linewidth}{!}{%
 \begin{tabular}{@{}l*{9}{c}@{}}
 \toprule
 & \multicolumn{3}{c}{\textbf{ALFWorld}} & \multicolumn{3}{c}{\textbf{SearchQA}} & \multicolumn{3}{c}{\textbf{SpreadsheetBench}} \\
 \cmidrule(lr){2-4}\cmidrule(lr){5-7}\cmidrule(l){8-10}
 \textbf{Method} & \textbf{Qwen3-32B} & \textbf{Qwen3.5-4B} & \textbf{Opus-4.6} & \textbf{Qwen3-32B} & \textbf{Qwen3.5-4B} & \textbf{Opus-4.6} & \textbf{Qwen3-32B} & \textbf{Qwen3.5-4B} & \textbf{Opus-4.6} \\
 \midrule
 \textbf{\OURS{}} & \makebox[2.3em][r]{\textbf{86.6}}\makebox[2.2em][l]{} & \makebox[2.3em][r]{\textbf{88.1}}\makebox[2.2em][l]{} & \makebox[2.3em][r]{\textbf{95.5}}\makebox[2.2em][l]{} & \makebox[2.3em][r]{\textbf{81.0}}\makebox[2.2em][l]{} & \makebox[2.3em][r]{\textbf{81.1}}\makebox[2.2em][l]{} & \makebox[2.3em][r]{\textbf{88.1}}\makebox[2.2em][l]{} & \makebox[2.3em][r]{\textbf{42.9}}\makebox[2.2em][l]{} & \makebox[2.3em][r]{\textbf{28.9}}\makebox[2.2em][l]{} & \makebox[2.3em][r]{\textbf{80.7}}\makebox[2.2em][l]{} \\
 One rollout & \cellcolor[HTML]{E9B6AB}\makebox[2.3em][r]{\textcolor{black}{76.9}}\makebox[2.2em][l]{\raisebox{-0.25ex}{\scriptsize\textcolor{black!55}{-9.7}}} & \cellcolor[HTML]{EFCBC3}\makebox[2.3em][r]{\textcolor{black}{81.3}}\makebox[2.2em][l]{\raisebox{-0.25ex}{\scriptsize\textcolor{black!55}{-6.7}}} & \cellcolor[HTML]{FCFBF8}\makebox[2.3em][r]{\textcolor{black}{95.5}}\makebox[2.2em][l]{\raisebox{-0.25ex}{\scriptsize\textcolor{black!55}{+0.0}}} & \cellcolor[HTML]{F5E2DC}\makebox[2.3em][r]{\textcolor{black}{77.4}}\makebox[2.2em][l]{\raisebox{-0.25ex}{\scriptsize\textcolor{black!55}{-3.6}}} & \cellcolor[HTML]{F4E0DA}\makebox[2.3em][r]{\textcolor{black}{77.3}}\makebox[2.2em][l]{\raisebox{-0.25ex}{\scriptsize\textcolor{black!55}{-3.8}}} & \cellcolor[HTML]{F9F0EC}\makebox[2.3em][r]{\textcolor{black}{86.6}}\makebox[2.2em][l]{\raisebox{-0.25ex}{\scriptsize\textcolor{black!55}{-1.6}}} & \cellcolor[HTML]{F6E7E1}\makebox[2.3em][r]{\textcolor{black}{40.0}}\makebox[2.2em][l]{\raisebox{-0.25ex}{\scriptsize\textcolor{black!55}{-2.9}}} & \cellcolor[HTML]{F3DDD6}\makebox[2.3em][r]{\textcolor{black}{24.6}}\makebox[2.2em][l]{\raisebox{-0.25ex}{\scriptsize\textcolor{black!55}{-4.3}}} & \cellcolor[HTML]{EBBEB4}\makebox[2.3em][r]{\textcolor{black}{72.1}}\makebox[2.2em][l]{\raisebox{-0.25ex}{\scriptsize\textcolor{black!55}{-8.6}}} \\
 No history & \cellcolor[HTML]{EDC6BD}\makebox[2.3em][r]{\textcolor{black}{79.1}}\makebox[2.2em][l]{\raisebox{-0.25ex}{\scriptsize\textcolor{black!55}{-7.5}}} & \cellcolor[HTML]{F9F0EC}\makebox[2.3em][r]{\textcolor{black}{86.6}}\makebox[2.2em][l]{\raisebox{-0.25ex}{\scriptsize\textcolor{black!55}{-1.5}}} & \cellcolor[HTML]{F2D6CF}\makebox[2.3em][r]{\textcolor{black}{90.3}}\makebox[2.2em][l]{\raisebox{-0.25ex}{\scriptsize\textcolor{black!55}{-5.2}}} & \cellcolor[HTML]{F4DED7}\makebox[2.3em][r]{\textcolor{black}{76.9}}\makebox[2.2em][l]{\raisebox{-0.25ex}{\scriptsize\textcolor{black!55}{-4.1}}} & \cellcolor[HTML]{F1D2CB}\makebox[2.3em][r]{\textcolor{black}{75.4}}\makebox[2.2em][l]{\raisebox{-0.25ex}{\scriptsize\textcolor{black!55}{-5.7}}} & \cellcolor[HTML]{FBFBF8}\makebox[2.3em][r]{\textcolor{black}{88.3}}\makebox[2.2em][l]{\raisebox{-0.25ex}{\scriptsize\textcolor{black!55}{+0.2}}} & \cellcolor[HTML]{F3DDD6}\makebox[2.3em][r]{\textcolor{black}{38.6}}\makebox[2.2em][l]{\raisebox{-0.25ex}{\scriptsize\textcolor{black!55}{-4.3}}} & \cellcolor[HTML]{FBF6F2}\makebox[2.3em][r]{\textcolor{black}{28.2}}\makebox[2.2em][l]{\raisebox{-0.25ex}{\scriptsize\textcolor{black!55}{-0.7}}} & \cellcolor[HTML]{F6E7E1}\makebox[2.3em][r]{\textcolor{black}{77.9}}\makebox[2.2em][l]{\raisebox{-0.25ex}{\scriptsize\textcolor{black!55}{-2.9}}} \\
 No regression bound & \cellcolor[HTML]{F5E1DB}\makebox[2.3em][r]{\textcolor{black}{82.8}}\makebox[2.2em][l]{\raisebox{-0.25ex}{\scriptsize\textcolor{black!55}{-3.7}}} & \cellcolor[HTML]{EFCBC3}\makebox[2.3em][r]{\textcolor{black}{81.3}}\makebox[2.2em][l]{\raisebox{-0.25ex}{\scriptsize\textcolor{black!55}{-6.7}}} & \cellcolor[HTML]{F9F0EC}\makebox[2.3em][r]{\textcolor{black}{94.0}}\makebox[2.2em][l]{\raisebox{-0.25ex}{\scriptsize\textcolor{black!55}{-1.5}}} & \cellcolor[HTML]{EBBDB3}\makebox[2.3em][r]{\textcolor{black}{72.3}}\makebox[2.2em][l]{\raisebox{-0.25ex}{\scriptsize\textcolor{black!55}{-8.7}}} & \cellcolor[HTML]{EABCB2}\makebox[2.3em][r]{\textcolor{black}{72.3}}\makebox[2.2em][l]{\raisebox{-0.25ex}{\scriptsize\textcolor{black!55}{-8.9}}} & \cellcolor[HTML]{F8ECE7}\makebox[2.3em][r]{\textcolor{black}{86.0}}\makebox[2.2em][l]{\raisebox{-0.25ex}{\scriptsize\textcolor{black!55}{-2.1}}} & \cellcolor[HTML]{EEC8C0}\makebox[2.3em][r]{\textcolor{black}{35.7}}\makebox[2.2em][l]{\raisebox{-0.25ex}{\scriptsize\textcolor{black!55}{-7.2}}} & \cellcolor[HTML]{EEC8C0}\makebox[2.3em][r]{\textcolor{black}{21.8}}\makebox[2.2em][l]{\raisebox{-0.25ex}{\scriptsize\textcolor{black!55}{-7.1}}} & \cellcolor[HTML]{E8B4A9}\makebox[2.3em][r]{\textcolor{black}{69.3}}\makebox[2.2em][l]{\raisebox{-0.25ex}{\scriptsize\textcolor{black!55}{-11.4}}} \\
 No validation gate & \cellcolor[HTML]{E9B6AB}\makebox[2.3em][r]{\textcolor{black}{76.9}}\makebox[2.2em][l]{\raisebox{-0.25ex}{\scriptsize\textcolor{black!55}{-9.7}}} & \cellcolor[HTML]{EDC6BD}\makebox[2.3em][r]{\textcolor{black}{80.6}}\makebox[2.2em][l]{\raisebox{-0.25ex}{\scriptsize\textcolor{black!55}{-7.5}}} & \cellcolor[HTML]{F5E1DB}\makebox[2.3em][r]{\textcolor{black}{91.8}}\makebox[2.2em][l]{\raisebox{-0.25ex}{\scriptsize\textcolor{black!55}{-3.7}}} & \cellcolor[HTML]{F6E7E1}\makebox[2.3em][r]{\textcolor{black}{78.1}}\makebox[2.2em][l]{\raisebox{-0.25ex}{\scriptsize\textcolor{black!55}{-2.9}}} & \cellcolor[HTML]{F5E2DC}\makebox[2.3em][r]{\textcolor{black}{77.6}}\makebox[2.2em][l]{\raisebox{-0.25ex}{\scriptsize\textcolor{black!55}{-3.6}}} & \cellcolor[HTML]{ECF2F4}\makebox[2.3em][r]{\textcolor{black}{91.1}}\makebox[2.2em][l]{\raisebox{-0.25ex}{\scriptsize\textcolor{black!55}{+3.0}}} & \cellcolor[HTML]{E8B4A9}\makebox[2.3em][r]{\textcolor{black}{30.7}}\makebox[2.2em][l]{\raisebox{-0.25ex}{\scriptsize\textcolor{black!55}{-12.2}}} & \cellcolor[HTML]{F7E9E4}\makebox[2.3em][r]{\textcolor{black}{26.4}}\makebox[2.2em][l]{\raisebox{-0.25ex}{\scriptsize\textcolor{black!55}{-2.5}}} & \cellcolor[HTML]{EFCDC5}\makebox[2.3em][r]{\textcolor{black}{74.3}}\makebox[2.2em][l]{\raisebox{-0.25ex}{\scriptsize\textcolor{black!55}{-6.4}}} \\
 \bottomrule
 \end{tabular}%
 }
 \endgroup
\end{table}

\paragraph{Q3: Does learned routing matter beyond skill content?}
To isolate routing, we fix the final library and compare four deployment rules. \textit{Admission scopes} restores the scope recorded when each expert was admitted. \textit{Random router} selects an expert at random. \textit{Share skill only} omits experts. \textit{Concatenated skills} loads the share skill and every expert for each task. In \textit{Frozen scopes}, each new expert retains its admission scope throughout training, while only skill content is revised. \OURS{} leads random routing, share-skill-only execution, and frozen scopes in all nine settings, concatenation in eight with one tie, and admission scopes in seven (Table~\ref{tab:ablation_routing}). The largest losses occur when experts are selected at random or omitted. These results support task-aware expert selection and continued adaptation of expert scopes.


\begin{table}[H]
 \centering
 \setlength{\belowcaptionskip}{3pt}
 \caption{Expert routing ablation. Black values are benchmark scores (\%). Gray subscripts show percentage-point changes from \OURS{}. Blue/red shading marks gains/losses.}
 \label{tab:ablation_routing}
 \begingroup
 \footnotesize
 \setlength{\tabcolsep}{2.5pt}
 \renewcommand{\arraystretch}{1.12}
 \resizebox{\linewidth}{!}{%
 \begin{tabular}{@{}l*{9}{c}@{}}
 \toprule
 & \multicolumn{3}{c}{\textbf{ALFWorld}} & \multicolumn{3}{c}{\textbf{SearchQA}} & \multicolumn{3}{c}{\textbf{SpreadsheetBench}} \\
 \cmidrule(lr){2-4}\cmidrule(lr){5-7}\cmidrule(l){8-10}
 \textbf{Method} & \textbf{Qwen3-32B} & \textbf{Qwen3.5-4B} & \textbf{Opus-4.6} & \textbf{Qwen3-32B} & \textbf{Qwen3.5-4B} & \textbf{Opus-4.6} & \textbf{Qwen3-32B} & \textbf{Qwen3.5-4B} & \textbf{Opus-4.6} \\
 \midrule
 \textbf{\OURS{}} & \makebox[2.3em][r]{\textbf{86.6}}\makebox[2.2em][l]{} & \makebox[2.3em][r]{\textbf{88.1}}\makebox[2.2em][l]{} & \makebox[2.3em][r]{\textbf{95.5}}\makebox[2.2em][l]{} & \makebox[2.3em][r]{\textbf{81.0}}\makebox[2.2em][l]{} & \makebox[2.3em][r]{\textbf{81.1}}\makebox[2.2em][l]{} & \makebox[2.3em][r]{\textbf{88.1}}\makebox[2.2em][l]{} & \makebox[2.3em][r]{\textbf{42.9}}\makebox[2.2em][l]{} & \makebox[2.3em][r]{\textbf{28.9}}\makebox[2.2em][l]{} & \makebox[2.3em][r]{\textbf{80.7}}\makebox[2.2em][l]{} \\
 Frozen scopes & \cellcolor[HTML]{EFCBC3}\makebox[2.3em][r]{\textcolor{black}{79.9}}\makebox[2.2em][l]{\raisebox{-0.25ex}{\scriptsize\textcolor{black!55}{-6.7}}} & \cellcolor[HTML]{F0D1C9}\makebox[2.3em][r]{\textcolor{black}{82.1}}\makebox[2.2em][l]{\raisebox{-0.25ex}{\scriptsize\textcolor{black!55}{-6.0}}} & \cellcolor[HTML]{F9F0EC}\makebox[2.3em][r]{\textcolor{black}{94.0}}\makebox[2.2em][l]{\raisebox{-0.25ex}{\scriptsize\textcolor{black!55}{-1.5}}} & \cellcolor[HTML]{EFCBC3}\makebox[2.3em][r]{\textcolor{black}{74.3}}\makebox[2.2em][l]{\raisebox{-0.25ex}{\scriptsize\textcolor{black!55}{-6.7}}} & \cellcolor[HTML]{F8ECE7}\makebox[2.3em][r]{\textcolor{black}{79.0}}\makebox[2.2em][l]{\raisebox{-0.25ex}{\scriptsize\textcolor{black!55}{-2.1}}} & \cellcolor[HTML]{F6E7E1}\makebox[2.3em][r]{\textcolor{black}{85.3}}\makebox[2.2em][l]{\raisebox{-0.25ex}{\scriptsize\textcolor{black!55}{-2.9}}} & \cellcolor[HTML]{EEC8C0}\makebox[2.3em][r]{\textcolor{black}{35.7}}\makebox[2.2em][l]{\raisebox{-0.25ex}{\scriptsize\textcolor{black!55}{-7.2}}} & \cellcolor[HTML]{F3DAD3}\makebox[2.3em][r]{\textcolor{black}{24.3}}\makebox[2.2em][l]{\raisebox{-0.25ex}{\scriptsize\textcolor{black!55}{-4.6}}} & \cellcolor[HTML]{E8B4A9}\makebox[2.3em][r]{\textcolor{black}{69.3}}\makebox[2.2em][l]{\raisebox{-0.25ex}{\scriptsize\textcolor{black!55}{-11.4}}} \\
 Admission scopes & \cellcolor[HTML]{F9F0EC}\makebox[2.3em][r]{\textcolor{black}{85.1}}\makebox[2.2em][l]{\raisebox{-0.25ex}{\scriptsize\textcolor{black!55}{-1.5}}} & \cellcolor[HTML]{F9F0EC}\makebox[2.3em][r]{\textcolor{black}{86.6}}\makebox[2.2em][l]{\raisebox{-0.25ex}{\scriptsize\textcolor{black!55}{-1.5}}} & \cellcolor[HTML]{F4F6F6}\makebox[2.3em][r]{\textcolor{black}{97.0}}\makebox[2.2em][l]{\raisebox{-0.25ex}{\scriptsize\textcolor{black!55}{+1.5}}} & \cellcolor[HTML]{F7EBE6}\makebox[2.3em][r]{\textcolor{black}{78.7}}\makebox[2.2em][l]{\raisebox{-0.25ex}{\scriptsize\textcolor{black!55}{-2.3}}} & \cellcolor[HTML]{F4F7F6}\makebox[2.3em][r]{\textcolor{black}{82.6}}\makebox[2.2em][l]{\raisebox{-0.25ex}{\scriptsize\textcolor{black!55}{+1.4}}} & \cellcolor[HTML]{F5E3DD}\makebox[2.3em][r]{\textcolor{black}{84.7}}\makebox[2.2em][l]{\raisebox{-0.25ex}{\scriptsize\textcolor{black!55}{-3.4}}} & \cellcolor[HTML]{F8EEEA}\makebox[2.3em][r]{\textcolor{black}{41.1}}\makebox[2.2em][l]{\raisebox{-0.25ex}{\scriptsize\textcolor{black!55}{-1.8}}} & \cellcolor[HTML]{F7E9E4}\makebox[2.3em][r]{\textcolor{black}{26.4}}\makebox[2.2em][l]{\raisebox{-0.25ex}{\scriptsize\textcolor{black!55}{-2.5}}} & \cellcolor[HTML]{FBF6F2}\makebox[2.3em][r]{\textcolor{black}{80.0}}\makebox[2.2em][l]{\raisebox{-0.25ex}{\scriptsize\textcolor{black!55}{-0.7}}} \\
 Concatenated skills & \cellcolor[HTML]{F8EBE6}\makebox[2.3em][r]{\textcolor{black}{84.3}}\makebox[2.2em][l]{\raisebox{-0.25ex}{\scriptsize\textcolor{black!55}{-2.2}}} & \cellcolor[HTML]{F5E1DB}\makebox[2.3em][r]{\textcolor{black}{84.3}}\makebox[2.2em][l]{\raisebox{-0.25ex}{\scriptsize\textcolor{black!55}{-3.7}}} & \cellcolor[HTML]{FCFBF8}\makebox[2.3em][r]{\textcolor{black}{95.5}}\makebox[2.2em][l]{\raisebox{-0.25ex}{\scriptsize\textcolor{black!55}{+0.0}}} & \cellcolor[HTML]{F4DFD8}\makebox[2.3em][r]{\textcolor{black}{77.0}}\makebox[2.2em][l]{\raisebox{-0.25ex}{\scriptsize\textcolor{black!55}{-4.0}}} & \cellcolor[HTML]{EDC4BB}\makebox[2.3em][r]{\textcolor{black}{73.4}}\makebox[2.2em][l]{\raisebox{-0.25ex}{\scriptsize\textcolor{black!55}{-7.7}}} & \cellcolor[HTML]{F7E9E4}\makebox[2.3em][r]{\textcolor{black}{85.6}}\makebox[2.2em][l]{\raisebox{-0.25ex}{\scriptsize\textcolor{black!55}{-2.6}}} & \cellcolor[HTML]{EFCDC5}\makebox[2.3em][r]{\textcolor{black}{36.4}}\makebox[2.2em][l]{\raisebox{-0.25ex}{\scriptsize\textcolor{black!55}{-6.4}}} & \cellcolor[HTML]{F4DFD9}\makebox[2.3em][r]{\textcolor{black}{25.0}}\makebox[2.2em][l]{\raisebox{-0.25ex}{\scriptsize\textcolor{black!55}{-3.9}}} & \cellcolor[HTML]{F2D7D0}\makebox[2.3em][r]{\textcolor{black}{75.7}}\makebox[2.2em][l]{\raisebox{-0.25ex}{\scriptsize\textcolor{black!55}{-5.0}}} \\
 Share skill only & \cellcolor[HTML]{ECC1B7}\makebox[2.3em][r]{\textcolor{black}{78.4}}\makebox[2.2em][l]{\raisebox{-0.25ex}{\scriptsize\textcolor{black!55}{-8.2}}} & \cellcolor[HTML]{F2D6CF}\makebox[2.3em][r]{\textcolor{black}{82.8}}\makebox[2.2em][l]{\raisebox{-0.25ex}{\scriptsize\textcolor{black!55}{-5.2}}} & \cellcolor[HTML]{F6E6E0}\makebox[2.3em][r]{\textcolor{black}{92.5}}\makebox[2.2em][l]{\raisebox{-0.25ex}{\scriptsize\textcolor{black!55}{-3.0}}} & \cellcolor[HTML]{EABBB1}\makebox[2.3em][r]{\textcolor{black}{72.0}}\makebox[2.2em][l]{\raisebox{-0.25ex}{\scriptsize\textcolor{black!55}{-9.0}}} & \cellcolor[HTML]{E8B4A9}\makebox[2.3em][r]{\textcolor{black}{68.7}}\makebox[2.2em][l]{\raisebox{-0.25ex}{\scriptsize\textcolor{black!55}{-12.4}}} & \cellcolor[HTML]{EECAC2}\makebox[2.3em][r]{\textcolor{black}{81.3}}\makebox[2.2em][l]{\raisebox{-0.25ex}{\scriptsize\textcolor{black!55}{-6.9}}} & \cellcolor[HTML]{F4DFD9}\makebox[2.3em][r]{\textcolor{black}{38.9}}\makebox[2.2em][l]{\raisebox{-0.25ex}{\scriptsize\textcolor{black!55}{-3.9}}} & \cellcolor[HTML]{F0D0C8}\makebox[2.3em][r]{\textcolor{black}{22.9}}\makebox[2.2em][l]{\raisebox{-0.25ex}{\scriptsize\textcolor{black!55}{-6.1}}} & \cellcolor[HTML]{E9B9AF}\makebox[2.3em][r]{\textcolor{black}{71.4}}\makebox[2.2em][l]{\raisebox{-0.25ex}{\scriptsize\textcolor{black!55}{-9.3}}} \\
 Random router & \cellcolor[HTML]{EABBB1}\makebox[2.3em][r]{\textcolor{black}{77.6}}\makebox[2.2em][l]{\raisebox{-0.25ex}{\scriptsize\textcolor{black!55}{-9.0}}} & \cellcolor[HTML]{EABBB1}\makebox[2.3em][r]{\textcolor{black}{79.1}}\makebox[2.2em][l]{\raisebox{-0.25ex}{\scriptsize\textcolor{black!55}{-9.0}}} & \cellcolor[HTML]{EDC6BD}\makebox[2.3em][r]{\textcolor{black}{88.1}}\makebox[2.2em][l]{\raisebox{-0.25ex}{\scriptsize\textcolor{black!55}{-7.5}}} & \cellcolor[HTML]{ECC2B9}\makebox[2.3em][r]{\textcolor{black}{73.0}}\makebox[2.2em][l]{\raisebox{-0.25ex}{\scriptsize\textcolor{black!55}{-8.0}}} & \cellcolor[HTML]{E8B4A9}\makebox[2.3em][r]{\textcolor{black}{67.0}}\makebox[2.2em][l]{\raisebox{-0.25ex}{\scriptsize\textcolor{black!55}{-14.1}}} & \cellcolor[HTML]{F1D3CC}\makebox[2.3em][r]{\textcolor{black}{82.6}}\makebox[2.2em][l]{\raisebox{-0.25ex}{\scriptsize\textcolor{black!55}{-5.6}}} & \cellcolor[HTML]{F1D5CE}\makebox[2.3em][r]{\textcolor{black}{37.5}}\makebox[2.2em][l]{\raisebox{-0.25ex}{\scriptsize\textcolor{black!55}{-5.4}}} & \cellcolor[HTML]{EABCB1}\makebox[2.3em][r]{\textcolor{black}{20.0}}\makebox[2.2em][l]{\raisebox{-0.25ex}{\scriptsize\textcolor{black!55}{-8.9}}} & \cellcolor[HTML]{E8B4A9}\makebox[2.3em][r]{\textcolor{black}{64.3}}\makebox[2.2em][l]{\raisebox{-0.25ex}{\scriptsize\textcolor{black!55}{-16.4}}} \\
 \bottomrule
 \end{tabular}%
 }
 \endgroup
\end{table}

\begin{figure}[!t]
 \centering
 \begin{minipage}[t]{0.48\linewidth}
  \centering
  \includegraphics[width=\linewidth]{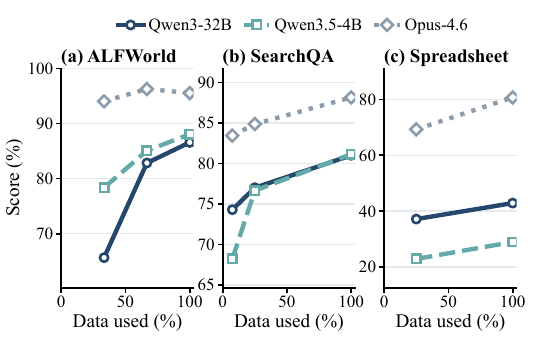}\vspace{-1em}
  \caption{Test-set scores by training-data fraction. Panels (a)--(c) show ALFWorld, SearchQA, and SpreadsheetBench, respectively.}
  \label{fig:experiment_data_efficiency}
 \end{minipage}\hfill
 \begin{minipage}[t]{0.48\linewidth}
  \centering
  \includegraphics[width=\linewidth]{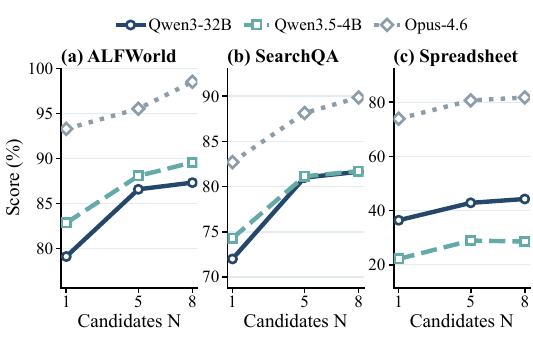}\vspace{-1em}
  \caption{Test-set scores by candidate count ($N$). Panels (a)--(c) show ALFWorld, SearchQA, and SpreadsheetBench.}
  \label{fig:experiment_candidate_count}
 \end{minipage}
\end{figure}

\paragraph{Q4: How do training data and unseen task families affect performance?}
We train on increasing fractions of the largest tested task subset per benchmark and evaluate each run on the same test set (Figure~\ref{fig:experiment_data_efficiency}). The largest subset outperforms the smallest in all nine benchmark--student settings. On SpreadsheetBench, moving from 25\% to 100\% improves Qwen3.5-4B by 6.0 points and Claude-Opus-4.6 by 11.4 points. We also exclude an entire ALFWorld task family during training and evaluate it afterward (Table~\ref{tab:appendix_e5_family}). All three students retain substantial success on both excluded families. These results show that greater data use generally improves test-set performance and that learned procedures remain useful on unseen task families without family-specific training examples.

\paragraph{Q5: How many candidate drafts are needed?}
We test $N\in\{1,5,8\}$ candidate drafts per target (Equation~\ref{eq:descent}), with $N=5$ as the default and other settings fixed (Figure~\ref{fig:experiment_candidate_count}). Increasing $N$ from one to five improves all nine benchmark--student settings by 6.2 points on average. At $N=8$, the average increases by another 1.1 points. Comparing several drafts, rather than relying on one proposal, therefore accounts for most of the benefit of deliberative skill revision.

\section{Conclusion}
In this work, we introduce \OURS{}, a mixture-of-skills framework that jointly optimizes a share skill, task-family experts, and their router. Each CreAGD optimization step follows five stages. First, \textit{forward execution and credit assignment} traces failures to their source. Second, \textit{textual and routing gradient descent} proposes revisions. Third, \textit{scope-conditioned paired evaluation} measures their effects. Fourth, \textit{history-calibrated step acceptance} selects updates. Finally, \textit{optimizer memory accumulation} guides later revisions. Across five benchmarks and five student models, \OURS{} improves average performance by 14.9--23.8 points over no skill and surpasses the strongest baseline. Ablations further validate the roles of its main components in these gains. These results highlight that evolving mixture-of-skills yields more adaptable agents.

\subsection*{AI use statement}
In this work, we used generative AI tools to help plan experimental comparisons and interpret and check results. We also used LLMs as the teacher, optimizer, router, and student agents in our experiments. We did not use generative AI to generate synthetic datasets or develop theoretical proofs, as these activities are not part of this study. Additionally, we used generative AI to polish author-written drafts, assist with subsequent manuscript revisions, and prepare figures and tables. We checked numerical claims, figures, and analyses against the experimental records, reviewed all AI-assisted work, and take responsibility for the final text, claims, and artifacts of this paper.

\subsection*{Ethics statement}
We study LLM-agent skill optimization using existing benchmarks. We did not recruit human participants or collect sensitive personal data. We are not aware of ethical concerns specific to this study beyond the general risks associated with deploying LLM agents.

\subsection*{Reproducibility statement}
We describe the method and experimental protocol in Sections~\ref{sec:optimization} and~\ref{sec:experimental_setup}. Appendices~\ref{app:experimental_details} and~\ref{app:experiments} document benchmark splits, hyperparameters, and additional results. We will release the implementation and experiment scripts in a code repository to support reproduction of our results.

\bibliography{iclr2027_conference}
\bibliographystyle{iclr2027_conference}
\clearpage
\appendix
\setcounter{table}{0}
\setcounter{figure}{0}
\setcounter{section}{0}
\setcounter{equation}{0}
\newcounter{promptcounter}
\setcounter{promptcounter}{0}
\renewcommand{\thetable}{A\arabic{table}}
\renewcommand{\thefigure}{A\arabic{figure}}
\renewcommand{\thesection}{A\arabic{section}}
\renewcommand{\theequation}{A\arabic{equation}}
\renewcommand{\theHtable}{appendix.A\arabic{table}}
\renewcommand{\theHfigure}{appendix.A\arabic{figure}}
\renewcommand{\theHsection}{appendix.A\arabic{section}}
\renewcommand{\theHequation}{appendix.A\arabic{equation}}
\raggedbottom
\section{Limitations}
\label{app:limitations}
While \OURS{} performs well across diverse agent tasks, several limitations remain. First, generating and validating multiple candidate skills requires repeated LLM calls and agent rollouts, increasing training time and token cost. More efficient candidate screening could reduce this overhead. Second, expert creation and routing rely on a partition of training tasks into families. When task boundaries overlap or shift, the resulting scopes may not identify where specialized guidance is needed. Third, a stronger teacher model diagnoses failures and proposes revisions. Errors in these diagnoses can propagate into skill updates, motivating further study of sensitivity to the teacher. Finally, our capacity study tests at most $K_{\max}=12$ experts. Larger libraries may make routing and expert maintenance more difficult, so their scalability remains to be evaluated.

\section{Experimental Details}
\label{app:experimental_details}

\subsection{Benchmarks and Data Splits}
\label{app:benchmark_splits}
The five benchmarks cover different task interfaces and evaluation objectives. ALFWorld~\citep{shridhar2021alfworld} presents household goals in a text environment, where an agent must complete a sequence of actions to succeed. We train on 90 tasks, use 24 tasks for validation, and report success on all 134 tasks in the unseen test split. ScienceWorld~\citep{wang2022scienceworld} requires interactive scientific experiments in a simulated environment. We use 60 training tasks and 18 validation tasks, and report the official best score on its 149-task test split. This progress-based score is reported separately from the binary task-success metric used for ALFWorld.

SearchQA~\citep{dunn2017searchqa} provides fixed retrieved snippets with each question and evaluates the resulting answer by exact match. SpreadsheetBench~\citep{ma2024spreadsheetbench} pairs natural-language instructions with real workbooks. A task counts as successful only when the modified workbook passes its checks. SkillLearnBench~\citep{zhong2026skilllearnbench} contains verifiable real-world tasks that test whether learned procedures improve execution. Table~\ref{tab:appendix_splits} lists the train, validation, and test sizes and primary metrics for all five benchmarks. All methods use the same test set. Training tasks support skill updates, validation tasks guide candidate screening and library selection, and test tasks are used only for final evaluation.

\begin{table}[H]
 \centering
 \caption{Benchmark splits and primary evaluation metrics. ALFWorld uses the unseen test split, and ScienceWorld uses its test split.}
 \label{tab:appendix_splits}
 \small
 \setlength{\tabcolsep}{5pt}
 \begin{tabular*}{\linewidth}{@{\extracolsep{\fill}}lrrrl@{}}
 \toprule
 Benchmark & Train & Validation & Test & Metric \\
 \midrule
 ALFWorld & 90 & 24 & 134 & Task success \\
 ScienceWorld & 60 & 18 & 149 & Official best score \\
 SearchQA & 400 & 200 & 1,400 & Exact match \\
 SpreadsheetBench & 80 & 40 & 280 & Hard accuracy \\
 SkillLearnBench & 51 & 18 & 26 & Task success \\
 \bottomrule
 \end{tabular*}
\end{table}

\subsection{Hyperparameter Settings}
\label{app:hyperparameters}
Table~\ref{tab:appendix_hyperparameters} summarizes the settings most relevant to agent execution and skill optimization. Student models remain frozen during skill learning and use the same decoding temperature for training rollouts and final evaluation. The teacher uses a higher temperature when proposing revisions, whereas student execution, routing, and paired evaluation use the lower temperature shown in the table. Execution budgets follow the task interface: ALFWorld and ScienceWorld require multiple environment actions, SearchQA uses a single answer, and SpreadsheetBench permits multi-turn code generation and workbook editing before the resulting file is checked.

Skill optimization processes training tasks in batches of 32, with $N=5$ candidate drafts proposed for each update target by default. Candidate and original systems are compared over three paired trials before history-calibrated checks and held-out validation determine acceptance. ALFWorld and ScienceWorld train for up to six epochs, while SearchQA and SpreadsheetBench train for up to four. Expert capacity also varies by benchmark to accommodate differences in task-family structure. Together, these settings specify the budgets for candidate search, validation, and final evaluation.

\begin{table}[H]
 \centering
 \caption{Selected execution and optimization hyperparameters. Benchmark-specific values are ordered as ALFWorld / ScienceWorld / SearchQA / SpreadsheetBench.}
 \label{tab:appendix_hyperparameters}
 \small
 \setlength{\tabcolsep}{5pt}
 \begin{tabular*}{\linewidth}{@{\extracolsep{\fill}}p{0.39\linewidth}p{0.20\linewidth}p{0.35\linewidth}@{}}
 \toprule
 Hyperparameter & Value & Notes \\
 \midrule
 \multicolumn{3}{@{}l}{\textit{LLM decoding}} \\
 Student temperature & 0.1 & Training and final evaluation \\
 Teacher temperature & 1.0 & Skill proposal and revision \\
 Router and gate temperature & 0.1 & Routing and paired evaluation \\
 Maximum output tokens & 16,384 & Student and teacher calls \\
 Random seed & 42 & Training and evaluation \\
 \midrule
 \multicolumn{3}{@{}l}{\textit{Agent execution}} \\
 Maximum execution steps & 50 / 40 / 1 / 30 & Benchmark-dependent agent budget \\
 \midrule
 \multicolumn{3}{@{}l}{\textit{Skill optimization}} \\
 Training epochs & 6 / 6 / 4 / 4 & Full training runs \\
 Batch size & 32 & Tasks per optimization batch \\
 Candidate drafts $N$ & 5 & Per skill-update target \\
 Paired gate trials & 3 & Candidate comparison \\
 Expert capacity $K_{\max}$ & 7 / 11 / 6 / 4 & Benchmark-specific expert limit \\
 \bottomrule
 \end{tabular*}
\end{table}

\section{Additional Experimental Results}
\label{app:experiments}

In this section, we report additional experiments and more detailed ablations of the components of \OURS{}. Unless stated otherwise, each comparison uses the test set for the same benchmark and student while retaining the execution and evaluation conditions of the full method.

\subsection{Skill Library Organization}
To assess skill-library organization, we train three alternatives from scratch: (1) \textbf{Single skill} optimizes one skill document for all tasks; (2) \textbf{Experts only} trains routed expert skills without a share skill; and (3) \textbf{Static partition} optimizes one skill per training family and assigns tasks by a fixed, description-based router. Table~\ref{tab:appendix_e1} reports test-set scores for three benchmarks and three students.
\begin{table}[h]
 \centering
 \caption{Skill-library organization: test-set scores (\%).}
 \label{tab:appendix_e1}
 \small
 \setlength{\tabcolsep}{5pt}
 \begin{tabular}{@{}llrrr@{}}
 \toprule
 Benchmark & Variant & Qwen3-32B & Qwen3.5-4B & Claude-Opus-4.6 \\
 \midrule
 \rowcolor{blue!7}
 ALFWorld & \textbf{Full} & \textbf{86.57} & \textbf{88.06} & \textbf{95.52} \\
  & \textbf{Single skill} & 76.12 & 79.10 & 89.55 \\
  & \textbf{Experts only} & 78.36 & 81.34 & 93.28 \\
  & \textbf{Static partition} & 83.58 & 82.84 & 94.78 \\
 \midrule
 \rowcolor{blue!7}
 SearchQA & \textbf{Full} & \textbf{81.00} & \textbf{81.14} & \textbf{88.14} \\
  & \textbf{Single skill} & 79.93 & 73.57 & 85.29 \\
  & \textbf{Experts only} & 78.86 & 79.14 & 85.71 \\
  & \textbf{Static partition} & 80.07 & 79.57 & 86.86 \\
 \midrule
 \rowcolor{blue!7}
 Spreadsheet & \textbf{Full} & \textbf{42.86} & \textbf{28.93} & \textbf{80.71} \\
  & \textbf{Single skill} & 37.50 & 18.21 & 72.50 \\
  & \textbf{Experts only} & 34.64 & 22.86 & 76.43 \\
  & \textbf{Static partition} & 43.93 & 25.71 & 81.43 \\
 \bottomrule
 \end{tabular}
\end{table}
\OURS{} exceeds \textbf{Single skill} and \textbf{Experts only} in all nine benchmark--student settings, by 6.8 and 4.7 percentage points on average, respectively. It also surpasses \textbf{Static partition} in seven settings and leads all three alternatives throughout ALFWorld and SearchQA. These results indicate that jointly optimizing the share skill, expert skills, and task assignments improves performance more consistently than the simpler library organizations.

\subsection{Interference from Accepted Skill Updates}
To examine whether accepted updates disrupt previously solved tasks, we audit expert-skill and share-skill edits on ALFWorld. For expert edits, we count both stable regressions, in which the previous version succeeds and the revised version fails across repeated trials, and any-trial regressions on previously stable tasks. For each share-skill edit, we compare family-level success under the original and revised systems on $Q=\mathcal D_{\mathrm{batch},t}$, the paired-evaluation batch spanning all six families. Table~\ref{tab:appendix_e2} separates update-level from family-level measurements.
\begin{table}[H]
 \centering
 \caption{Effects of accepted ALFWorld skill updates. Expert regressions are counted across accepted edits. Share-skill regressions are counted across family evaluations.}
 \label{tab:appendix_e2}
 \small
 \begin{tabular}{@{}lrrl@{}}
 \toprule
 Update lane & Accepted edits & Audit units & Observed regression \\
 \midrule
 Expert skill & 27 & 27 edits & 0 stable, 8 any-trial \\
 Share skill & 7 & 42 family checks & 2 one-task regressions \\
 \bottomrule
 \end{tabular}
\end{table}
None of the 27 accepted expert edits caused a stable regression, although eight showed an any-trial regression. Of the 42 family-level comparisons for seven accepted share-skill edits, only two were negative, each by one successful task. The acceptance checks rejected candidates with larger family-level losses. Together, these results show that scoped expert edits preserve stable behavior and that regression checks limit interference from global share-skill revisions.

\subsection{Reliability of Update Acceptance}
To measure the contribution of the acceptance checks in Section~\ref{sec:optimization}, we retrain after changing one check at a time. \textbf{One rollout} replaces repeated evaluation with a single run. \textbf{No history} removes historical calibration. \textbf{No regression bound} removes the stable-regression criterion. \textbf{No validation gate} removes held-out validation. Table~\ref{tab:appendix_e3} reports the final test-set scores.
\begin{table}[H]
 \centering
 \caption{Update-acceptance ablations: test-set scores (\%).}
 \label{tab:appendix_e3}
 \small
 \setlength{\tabcolsep}{5pt}
 \begin{tabular}{@{}llrrr@{}}
 \toprule
 Benchmark & Variant & Qwen3-32B & Qwen3.5-4B & Claude-Opus-4.6 \\
 \midrule
 \rowcolor{blue!7}
 ALFWorld & \textbf{Full} & \textbf{86.57} & \textbf{88.06} & \textbf{95.52} \\
  & \textbf{One rollout} & 76.87 & 81.34 & 95.52 \\
  & \textbf{No history} & 79.10 & 86.57 & 90.30 \\
  & \textbf{No regression bound} & 82.84 & 81.34 & 94.03 \\
  & \textbf{No validation gate} & 76.87 & 80.60 & 91.79 \\
 \midrule
 \rowcolor{blue!7}
 SearchQA & \textbf{Full} & \textbf{81.00} & \textbf{81.14} & \textbf{88.14} \\
  & \textbf{One rollout} & 77.43 & 77.29 & 86.57 \\
  & \textbf{No history} & 76.86 & 75.43 & 88.29 \\
  & \textbf{No regression bound} & 72.29 & 72.29 & 86.00 \\
  & \textbf{No validation gate} & 78.14 & 77.57 & 91.14 \\
 \midrule
 \rowcolor{blue!7}
 Spreadsheet & \textbf{Full} & \textbf{42.86} & \textbf{28.93} & \textbf{80.71} \\
  & \textbf{One rollout} & 40.00 & 24.64 & 72.14 \\
  & \textbf{No history} & 38.57 & 28.21 & 77.86 \\
  & \textbf{No regression bound} & 35.71 & 21.79 & 69.29 \\
  & \textbf{No validation gate} & 30.71 & 26.43 & 74.29 \\
 \bottomrule
 \end{tabular}
\end{table}
Removing the regression bound lowers scores in all nine settings by 6.4 percentage points on average. Weakening or removing each of the other checks lowers scores in eight settings. The regression bound has the clearest effect, while repeated trials, historical calibration, and held-out validation also contribute consistently. The combined checks support reliable updates across tasks and students.

\subsection{The Role of Learned Routing}
To distinguish the effects of routing and skill text, we evaluate five alternatives. \textbf{Frozen scopes} fixes the scope of each new expert at admission while continuing to update skill text during training. The other four hold the final skill texts fixed: \textbf{Admission scopes} restores the scopes recorded when experts were admitted; \textbf{Random router} selects an expert at random; \textbf{Share skill only} omits experts; and \textbf{Concatenated skills} loads the share skill and every expert for each task. Together, these settings test scope adaptation during training and expert selection at deployment.
\begin{table}[!t]
 \centering
 \caption{Routing interventions: test-set scores (\%).}
 \label{tab:appendix_e4}
 \small
 \setlength{\tabcolsep}{5pt}
 \begin{tabular}{@{}llrrr@{}}
 \toprule
 Benchmark & Variant & Qwen3-32B & Qwen3.5-4B & Claude-Opus-4.6 \\
 \midrule
 \rowcolor{blue!7}
 ALFWorld & \textbf{Full} & \textbf{86.57} & \textbf{88.06} & \textbf{95.52} \\
  & \textbf{Frozen scopes} & 79.85 & 82.09 & 94.03 \\
  & \textbf{Random router} & 77.61 & 79.10 & 88.06 \\
  & \textbf{Share skill only} & 78.36 & 82.84 & 92.54 \\
  & \textbf{Concatenated skills} & 84.33 & 84.33 & 95.52 \\
  & \textbf{Admission scopes} & 85.07 & 86.57 & 97.01 \\
 \midrule
 \rowcolor{blue!7}
 SearchQA & \textbf{Full} & \textbf{81.00} & \textbf{81.14} & \textbf{88.14} \\
  & \textbf{Frozen scopes} & 74.29 & 79.00 & 85.29 \\
  & \textbf{Random router} & 73.00 & 67.00 & 82.57 \\
  & \textbf{Share skill only} & 72.00 & 68.71 & 81.29 \\
  & \textbf{Concatenated skills} & 77.00 & 73.43 & 85.57 \\
  & \textbf{Admission scopes} & 78.71 & 82.57 & 84.71 \\
 \midrule
 \rowcolor{blue!7}
 Spreadsheet & \textbf{Full} & \textbf{42.86} & \textbf{28.93} & \textbf{80.71} \\
  & \textbf{Frozen scopes} & 35.71 & 24.29 & 69.29 \\
  & \textbf{Random router} & 37.50 & 20.00 & 64.29 \\
  & \textbf{Share skill only} & 38.93 & 22.86 & 71.43 \\
  & \textbf{Concatenated skills} & 36.43 & 25.00 & 75.71 \\
  & \textbf{Admission scopes} & 41.07 & 26.43 & 80.00 \\
 \bottomrule
 \end{tabular}
\end{table}
\OURS{} exceeds \textbf{Frozen scopes}, \textbf{Random router}, and \textbf{Share skill only} in all nine settings. It leads \textbf{Concatenated skills} in eight settings with one tie, and \textbf{Admission scopes} in seven. Random selection or omission of experts produces the largest average losses. These comparisons show the value of task-aware expert selection, while the lower scores under \textbf{Frozen scopes} also support scope adaptation during training.

\subsection{Training-Data Efficiency and Unseen-Family Generalization}
To assess data efficiency, we vary the fraction of training tasks while keeping validation and test sets fixed. For ALFWorld, 30, 60, and 90 tasks represent 33.3\%, 66.7\%, and 100\% of the largest subset tested. For SearchQA, 30, 100, and 400 tasks represent 7.5\%, 25\%, and 100\%. For SpreadsheetBench, 20 and 80 tasks represent 25\% and 100\%. We also test unseen-family generalization by excluding either the ALFWorld two-object or look family during training and evaluating on the excluded family. Tables~\ref{tab:appendix_e5} and~\ref{tab:appendix_e5_family} report the aggregate and family-specific scores.
\begin{table}[h]
 \centering
 \caption{Test-set scores (\%) by training-data fraction and ALFWorld family holdout. Fractions are relative to the largest training subset tested for each benchmark.}
 \label{tab:appendix_e5}
 \small
 \setlength{\tabcolsep}{5pt}
 \begin{tabular}{@{}llrrr@{}}
 \toprule
 Benchmark & Variant & Qwen3-32B & Qwen3.5-4B & Claude-Opus-4.6 \\
 \midrule
 ALFWorld & \textbf{33.3\% data} & 65.67 & 78.36 & 94.03 \\
  & \textbf{66.7\% data} & 82.84 & 85.07 & 96.27 \\
  & \textbf{100\% data} & 86.57 & 88.06 & 95.52 \\
  & \textbf{Hold out two-object} & 82.84 & 85.82 & 94.03 \\
  & \textbf{Hold out look} & 83.58 & 85.07 & 95.52 \\
 \midrule
 SearchQA & \textbf{7.5\% data} & 74.29 & 68.21 & 83.43 \\
  & \textbf{25\% data} & 77.00 & 76.64 & 84.86 \\
  & \textbf{100\% data} & 81.00 & 81.14 & 88.14 \\
 \midrule
 Spreadsheet & \textbf{25\% data} & 37.14 & 22.86 & 69.29 \\
  & \textbf{100\% data} & 42.86 & 28.93 & 80.71 \\
 \bottomrule
 \end{tabular}
\end{table}
\begin{table}[h]
 \centering
 \caption{ALFWorld success rates on task families excluded from training. Decimal values are successful unseen tasks divided by the family test size ($n=17$ for two-object and $n=18$ for look). Full includes the family during training, whereas Family holdout excludes it.}
 \label{tab:appendix_e5_family}
 \small
 \setlength{\tabcolsep}{8pt}
 \renewcommand{\arraystretch}{1.12}
 \begin{tabular}{@{}llccc@{}}
 \toprule
 Excluded family & Training & Qwen3-32B & Qwen3.5-4B & Claude-Opus-4.6 \\
 \midrule
 Two-object ($n=17$) & \textbf{Full} & 0.706 & 1.000 & 1.000 \\
  & \textbf{Family holdout} & 0.529 & 0.824 & 0.882 \\
 \addlinespace[4pt]
 Look ($n=18$) & \textbf{Full} & 1.000 & 0.889 & 1.000 \\
  & \textbf{Family holdout} & 0.889 & 0.722 & 1.000 \\
 \bottomrule
 \end{tabular}
\end{table}
The largest training subset outperforms the smallest in all nine benchmark--student settings, although the ALFWorld score of Claude-Opus-4.6 peaks at the intermediate fraction. When an entire ALFWorld family is excluded, success rates remain 0.529--0.882 on two-object and 0.722--1.000 on look (Table~\ref{tab:appendix_e5_family}). Students solve at most three fewer tasks than when the family is included in training. Thus, more data generally improves test-set performance, and procedures learned from other families remain useful on unseen tasks.

\subsection{Cross-Student Transfer of Skill Libraries}
To test whether a skill library transfers between student models, we train a separate source library for each of three students on each benchmark. At evaluation, the source skill texts and router scopes are frozen and used by a different target student without further training. We compare each transfer with both no-skill execution and the library trained directly for the target student, shown on the matrix diagonal of Table~\ref{tab:appendix_e6}. The SpreadsheetBench libraries use two training epochs.
\begin{table}[H]
 \centering
 \caption{Cross-student transfer: target test-set scores (\%). Rows identify the source library, and columns identify the executing student. Diagonal cells use libraries trained for the target student. Bold marks the highest score in each benchmark--target column. ALFWorld and SpreadsheetBench use hard accuracy ($n=134$ and $280$). SearchQA uses exact match ($n=1{,}400$).}
 \label{tab:appendix_e6}
 \small
 \renewcommand{\arraystretch}{1.12}
 \begin{tabular*}{\linewidth}{@{\extracolsep{\fill}}lccc@{}}
 \toprule
 Source library & Qwen3-32B & Qwen3.5-4B & Claude-Opus-4.6 \\
 \midrule
 \multicolumn{4}{@{}l}{\textit{ALFWorld}} \\
 \textbf{No skill} & 40.30 & 32.09 & 80.60 \\
 \textbf{Qwen3-32B} & \cellcolor{blue!7}\textbf{86.57} & 82.09 & 87.31 \\
 \textbf{Qwen3.5-4B} & 76.87 & \cellcolor{blue!7}\textbf{88.06} & 91.04 \\
 \textbf{Claude-Opus-4.6} & 80.60 & 64.93 & \cellcolor{blue!7}\textbf{95.52} \\
 \addlinespace[3pt]
 \cmidrule(lr){1-4}
 \multicolumn{4}{@{}l}{\textit{SearchQA}} \\
 \textbf{No skill} & 72.29 & 68.21 & 82.79 \\
 \textbf{Qwen3-32B} & \cellcolor{blue!7}\textbf{81.00} & 77.43 & 86.86 \\
 \textbf{Qwen3.5-4B} & 78.14 & \cellcolor{blue!7}\textbf{81.14} & 87.36 \\
 \textbf{Claude-Opus-4.6} & 75.14 & 72.43 & \cellcolor{blue!7}\textbf{88.14} \\
 \addlinespace[3pt]
 \cmidrule(lr){1-4}
 \multicolumn{4}{@{}l}{\textit{SpreadsheetBench}} \\
 \textbf{No skill} & 22.50 & 9.64 & 56.79 \\
 \textbf{Qwen3-32B} & \cellcolor{blue!7}\textbf{42.86} & 27.50 & 74.29 \\
 \textbf{Qwen3.5-4B} & 36.43 & \cellcolor{blue!7}\textbf{28.93} & 69.29 \\
 \textbf{Claude-Opus-4.6} & 39.29 & 23.57 & \cellcolor{blue!7}\textbf{80.71} \\
 \bottomrule
 \end{tabular*}
\end{table}
All 18 cross-student deployments improve on the No skill score of the target student, with gains of 2.85--50.00 percentage points. In every benchmark--target setting, the library trained directly for the target achieves the highest score, exceeding all 18 transferred alternatives. These results show that skill texts and routing scopes learned with one student remain useful to another without additional training, while target-specific optimization provides the strongest test-set performance.

\subsection{Expert Lifecycle and Optimizer Memory}
To test the contribution of library maintenance, we retrain after removing one mechanism at a time. \textbf{No retirement} keeps experts despite sustained degradation. \textbf{No draft reuse} permits retirement but does not reuse retired drafts in later proposals. \textbf{No rejected memory} retains other optimizer records but omits rejected candidates and their feedback (Table~\ref{tab:appendix_e7}).
\begin{table}[h]
 \centering
 \caption{Expert-lifecycle and memory ablations: test-set scores (\%).}
 \label{tab:appendix_e7}
 \small
 \setlength{\tabcolsep}{5pt}
 \begin{tabular}{@{}llrrr@{}}
 \toprule
 Benchmark & Variant & Qwen3-32B & Qwen3.5-4B & Claude-Opus-4.6 \\
 \midrule
 \rowcolor{blue!7}
 ALFWorld & \textbf{Full} & \textbf{86.57} & \textbf{88.06} & \textbf{95.52} \\
  & \textbf{No retirement} & 76.87 & 73.88 & 92.54 \\
  & \textbf{No draft reuse} & 80.60 & 81.34 & 91.04 \\
  & \textbf{No rejected memory} & 81.34 & 82.09 & 91.79 \\
 \midrule
 \rowcolor{blue!7}
 SearchQA & \textbf{Full} & \textbf{81.00} & \textbf{81.14} & \textbf{88.14} \\
  & \textbf{No retirement} & 70.57 & 73.50 & 83.29 \\
  & \textbf{No draft reuse} & 72.86 & 69.29 & 84.43 \\
  & \textbf{No rejected memory} & 74.29 & 72.29 & 84.86 \\
 \midrule
 \rowcolor{blue!7}
 Spreadsheet & \textbf{Full} & \textbf{42.86} & \textbf{28.93} & \textbf{80.71} \\
  & \textbf{No retirement} & 35.71 & 12.86 & 66.07 \\
  & \textbf{No draft reuse} & 33.57 & 23.21 & 72.14 \\
  & \textbf{No rejected memory} & 33.57 & 17.14 & 68.21 \\
 \bottomrule
 \end{tabular}
\end{table}
Removing any of the three mechanisms lowers scores in all nine settings. The full method leads \textbf{No retirement}, \textbf{No draft reuse}, and \textbf{No rejected memory} by 9.7, 7.2, and 7.5 percentage points on average, respectively. Retirement has the largest mean effect, while draft reuse and rejected-candidate memory also provide consistent gains. These results show that lifecycle maintenance and optimizer memory support sustained skill improvement.

\subsection{Contribution of Paired Teacher--Student Evidence}
To test whether task-matched teacher--student evidence improves skill revision, we vary the trajectories available to the optimizer. \textbf{Full} compares teacher successes with student executions on the same tasks. \textbf{Failures only} provides student failure trajectories, \textbf{Successes only} provides teacher success trajectories, and \textbf{Unpaired evidence} provides both types without task-level matching (Table~\ref{tab:appendix_e8}).
\begin{table}[H]
 \centering
 \caption{Teacher--student evidence ablations: test-set scores (\%).}
 \label{tab:appendix_e8}
 \small
 \setlength{\tabcolsep}{5pt}
 \begin{tabular}{@{}llrrr@{}}
 \toprule
 Benchmark & Variant & Qwen3-32B & Qwen3.5-4B & Claude-Opus-4.6 \\
 \midrule
 \rowcolor{blue!7}
 ALFWorld & \textbf{Full} & \textbf{86.57} & \textbf{88.06} & \textbf{95.52} \\
  & \textbf{Failures only} & 76.87 & 75.37 & 94.03 \\
  & \textbf{Successes only} & 78.36 & 81.34 & 92.54 \\
  & \textbf{Unpaired evidence} & 81.34 & 85.07 & 89.55 \\
 \midrule
 \rowcolor{blue!7}
 SearchQA & \textbf{Full} & \textbf{81.00} & \textbf{81.14} & \textbf{88.14} \\
  & \textbf{Failures only} & 73.71 & 76.29 & 83.29 \\
  & \textbf{Successes only} & 75.71 & 72.14 & 86.29 \\
  & \textbf{Unpaired evidence} & 73.57 & 75.57 & 80.86 \\
 \midrule
 \rowcolor{blue!7}
 Spreadsheet & \textbf{Full} & \textbf{42.86} & \textbf{28.93} & \textbf{80.71} \\
  & \textbf{Failures only} & 34.29 & 20.00 & 70.71 \\
  & \textbf{Successes only} & 38.57 & 22.86 & 71.43 \\
  & \textbf{Unpaired evidence} & 39.29 & 15.71 & 66.43 \\
 \bottomrule
 \end{tabular}
\end{table}
\textbf{Full} exceeds all three restricted-evidence variants in every setting, by 7.6, 6.0, and 7.3 percentage points on average relative to \textbf{Failures only}, \textbf{Successes only}, and \textbf{Unpaired evidence}. The consistent advantage over both single-source and unpaired evidence shows that matching teacher successes with student executions on the same task provides more useful information for skill revision.

\subsection{Candidate Count and Expert Capacity}
To assess candidate search and expert capacity, we vary one setting at a time. The number of drafts per update target is \textbf{$N=5$} by default, compared with \textbf{$N=1$} and \textbf{$N=8$}. The default expert limits for ALFWorld, SearchQA, and SpreadsheetBench are 7, 6, and 4, respectively. The capacity scan sets each limit to \textbf{$K_{\max}=12$} while retaining five drafts (Table~\ref{tab:appendix_e9}).
\begin{table}[h]
 \centering
 \caption{Candidate-count and expert-capacity scans: test-set scores (\%).}
 \label{tab:appendix_e9}
 \small
 \setlength{\tabcolsep}{5pt}
 \begin{tabular}{@{}llrrr@{}}
 \toprule
 Benchmark & Variant & Qwen3-32B & Qwen3.5-4B & Claude-Opus-4.6 \\
 \midrule
 \rowcolor{blue!7}
 ALFWorld & \textbf{Default ($N=5$, $K_{\max}=7$)} & \textbf{86.57} & \textbf{88.06} & \textbf{95.52} \\
  & \textbf{$N=1$} & 79.10 & 82.84 & 93.28 \\
  & \textbf{$N=8$} & 87.31 & 89.55 & 97.01 \\
  & \textbf{$K_{\max}=12$} & 87.31 & 88.06 & 97.76 \\
 \midrule
 \rowcolor{blue!7}
 SearchQA & \textbf{Default ($N=5$, $K_{\max}=6$)} & \textbf{81.00} & \textbf{81.14} & \textbf{88.14} \\
  & \textbf{$N=1$} & 72.00 & 74.29 & 82.71 \\
  & \textbf{$N=8$} & 81.64 & 81.71 & 89.86 \\
  & \textbf{$K_{\max}=12$} & 81.57 & 78.43 & 84.00 \\
 \midrule
 \rowcolor{blue!7}
 Spreadsheet & \textbf{Default ($N=5$, $K_{\max}=4$)} & \textbf{42.86} & \textbf{28.93} & \textbf{80.71} \\
  & \textbf{$N=1$} & 36.43 & 22.14 & 73.93 \\
  & \textbf{$N=8$} & 44.29 & 28.57 & 81.79 \\
  & \textbf{$K_{\max}=12$} & 43.21 & 26.79 & 83.21 \\
 \bottomrule
 \end{tabular}
\end{table}
Reducing the candidate count from five to one lowers all nine scores by 6.2 percentage points on average. Increasing it to eight improves eight settings and adds a further 1.1 points on average. Raising each expert-capacity limit to twelve yields five gains, three declines, and one tie, with a 0.3-point mean gain. Multiple candidate drafts drive the main improvement, while the small and uneven return from more experts suggests that the benchmark-specific default capacities are sufficient for these tasks.

\subsection{SearchQA Performance Across Learning Budgets}
\begin{figure}[H]
 \centering
 \begin{minipage}[t]{0.49\linewidth}
  \centering
  \includegraphics[width=\linewidth]{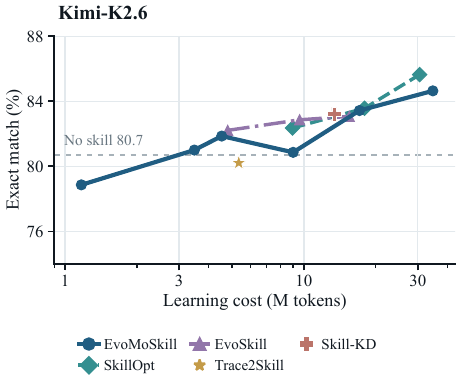}
 \end{minipage}\hfill
 \begin{minipage}[t]{0.49\linewidth}
  \centering
  \includegraphics[width=\linewidth]{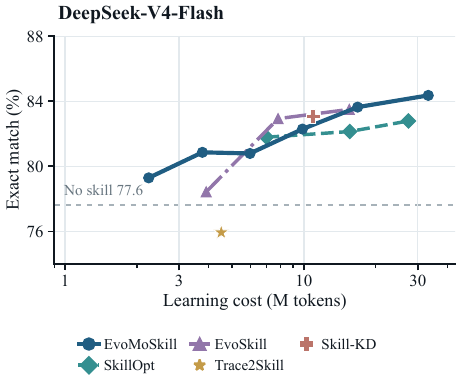}
 \end{minipage}
 \caption{SearchQA exact match versus learning cost for Kimi-K2.6 and DeepSeek-V4-Flash.}
 \label{fig:appendix_searchqa_budget}
\end{figure}

To assess learning-cost performance on question answering, we evaluate the same five skill methods on SearchQA with Kimi-K2.6 and DeepSeek-V4-Flash. Figure~\ref{fig:appendix_searchqa_budget} shows test-set exact match against cumulative learning tokens, with dashed No skill references.

Across both students, \OURS{} rises overall with more learning and ends above No skill. SkillOpt has the best final Kimi-K2.6 score, whereas \OURS{} leads on DeepSeek-V4-Flash. The curves show sustained gains from iterative skill optimization on SearchQA.
\end{document}